%% file: neurips_2023.tex
\pdfoutput=1

\documentclass{article}

\PassOptionsToPackage{numbers, compress}{natbib}

\usepackage[final]{neurips_2023}

\usepackage[utf8]{inputenc}
\usepackage{times}
\usepackage{latexsym}

\usepackage{hyperref}       
\usepackage{url}            
\usepackage{booktabs}       
\usepackage{amsfonts}       
\usepackage{nicefrac}       
\usepackage{microtype}      
\usepackage{xcolor, soul}   
\sethlcolor{lime}

\usepackage{enumitem}
\usepackage{latexsym}
\usepackage{adjustbox}
\usepackage[hang,flushmargin]{footmisc}
\usepackage{multirow}
\usepackage{comment}

\usepackage{amsmath}
\usepackage{tabularx}
\usepackage{algorithm}
\usepackage{algpseudocode}
\usepackage{wrapfig}
\usepackage{threeparttable}

\makeatletter
\newcommand{\multiline}[1]{%
  \begin{tabularx}{\dimexpr\linewidth-\ALG@thistlm}[t]{@{}X@{}}
    #1
  \end{tabularx}
  \vspace{-0.4em}
}
\makeatother

\DeclareMathOperator{\argmax}{argmax}

\input{sections/macros}

\title{%
    BERT Lost Patience \\Won't Be Robust to Adversarial Slowdown}

\author{
Zachary Coalson,
Gabriel Ritter,
Rakesh Bobba,
and Sanghyun Hong \\
Oregon State University \\
\texttt{\{coalsonz, ritterg, bobbar, hongsa\}@oregonstate.edu}
}

\newcommand{\topic}[1]{\noindent\textbf{#1}}

\newcommand{\SH}[1]{{\color{red}\hl{SH: #1}}}

\begin{document}

\maketitle

\input{sections/abstract}

\input{sections/intro}

\input{sections/related}

\input{sections/methods}

\input{sections/results}
\input{sections/discussion}

\input{sections/conclusion}

\input{sections/limitation}

\newpage

\input{sections/ack}

{
    \bibliographystyle{abbrvnat}
    \bibliography{bib/nlpslowdown}
}


\newpage
\appendix
\input{sections/appendix}

\end{document}

%% file: sections/macros.tex
\newcommand{\attname}{\textsc{Waffle}}

%% file: sections/abstract.tex
%
%
\begin{abstract}
In this paper, we systematically evaluate the robustness of
multi-exit language models against adversarial slowdown.
To \emph{audit} their robustness,
we design a slowdown attack
that generates natural adversarial text bypassing early-exit points.
We use the resulting \attname{} attack as a vehicle
to conduct a comprehensive evaluation of three
multi-exit mechanisms
with the GLUE benchmark against adversarial slowdown.
We then show our attack significantly reduces the computational savings 
provided by the three methods in both white-box and black-box settings.
The more complex a mechanism is,
the more vulnerable it is to adversarial slowdown.
We also perform a linguistic analysis of the perturbed text inputs,
identifying common perturbation patterns that our attack generates, 
and comparing them with standard adversarial text attacks.
Moreover, we show that 
adversarial training is ineffective in defeating our slowdown attack,
but input sanitization with a conversational model, e.g., ChatGPT,
can remove perturbations effectively.
This result suggests that future work is needed for
developing efficient yet robust multi-exit models.
Our code is available at: {\footnotesize\href{https://github.com/ztcoalson/WAFFLE}{https://github.com/ztcoalson/WAFFLE}}
\end{abstract}

%% file: sections/intro.tex
\section{Introduction}
\label{sec:intro}

%
A key factor behind the recent 
advances in natural language processing
is the \emph{scale} of language models 
pre-trained on a large corpus of data.
Compared to BERT~\cite{BERT} with 110 million parameters
that achieves the GLUE benchmark score~\cite{GLUE} 
of 81\% from three years ago,
T5~\cite{T5} improves the score to 90\%
with 100$\times$ more parameters.
However, pre-trained language models with this scale
typically require large memory and high computational costs 
to run 
inferences, making them challenging in scenarios
where latency and computations are limited.

%
To address this issue,
\emph{input-adaptive} multi-exit mechanisms~\citep{RTRJ:20, DeeBERT, PABEE, ELBERT:21, BERxIT:21, LeeBERT:21, PastFuture, PCEE:22} have been proposed.
By attaching internal classifiers (or early exits) 
to each intermediate layer of a pre-trained language model,
the resulting multi-exit language model 
utilizes these exits to stop its forward pass preemptively,
when the model is confident about its prediction at any exit point.
This prevents 
models
from spending excessive computation for ``easy" inputs,
where shallow models are sufficient for correct predictions,
and therefore reduces the post-training workloads while preserving accuracy.

%
In this work, we study the robustness of multi-exit language models 
to \emph{adversarial slowdown}.
Recent work~\cite{DeepSloth} showed that, 
against multi-exit models developed for computer vision tasks,
an adversary can craft human-imperceptible input perturbations 
to offset 
their 
computational savings. 
However, it has not yet been shown that
the input-adaptive 
methods proposed in language domains
are susceptible to such input perturbations.
It is also unknown 
why these perturbations cause slowdown and 
how similar they are to those generated by standard adversarial 
attacks.
Moreover, it is unclear if existing defenses, 
e.g., adversarial training~\cite{A2T},
proposed in the community
can mitigate slowdown attacks.

\begin{wrapfigure}{r}{0.5\linewidth} 
    \vspace{-1.2em}
    \includegraphics[width=1.0\linewidth]{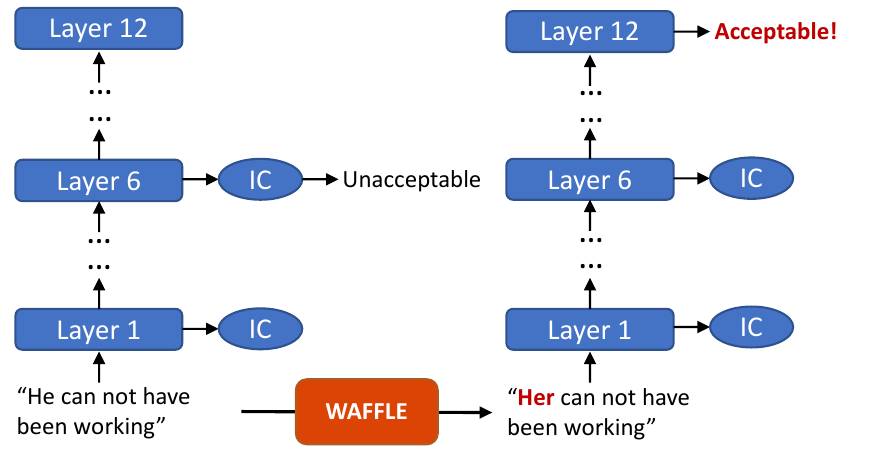}
    \caption{\textbf{Illustrating adversarial slowdown.} Replacing the word ``He" with ``Her" makes the resulting text input bypass all 11 ICs (internal classifiers) and leads to misclassification. The text is chosen from the Corpus of Linguistic Acceptability.}
    \label{fig:attack-overview}
    \vspace{-1.0em}
\end{wrapfigure}

%
\textbf{Our contributions.}
To bridge this gap, we \emph{first} develop \attname{},
a slowdown attack that generates natural adversarial text
that bypasses early-exits.
We illustrate how our attack works in Figure~\ref{fig:attack-overview}.
Based on our finding that
existing adversarial text attacks~\citep{TextFooler, A2T}
fail to cause 
significant slowdown, 
we design a novel objective function
that pushes a multi-exit model's predictions at its early-exits toward the uniform distribution.
\attname{} integrates this objective into existing 
attack algorithms.

\emph{Second}, we 
systematically evaluate the robustness of three 
early-exit mechanisms~\cite{DeeBERT, PABEE, PastFuture}
on the GLUE benchmark 
against adversarial slowdown.
We find that \attname{} significantly offsets
the computational savings provided by the mechanisms 
when each text input is individually subject to perturbations.
We also show that
methods offering more aggressive computational savings
are more vulnerable to our 
attack.

\emph{Third},
we show that \attname{} can be effective in black-box scenarios.
We demonstrate that our attack transfers,
i.e., the adversarial texts 
crafted with limited knowledge about the victim
cause slowdown across different models and multi-exit mechanisms.
We are also able to find universal slowdown triggers,
i.e., input-agnostic sequences of words
that reduces the computational savings of multi-exit language models
when attached to any text input from a dataset.

\emph{Fourth}, 
we conduct a linguistic analysis of the adversarial texts \attname{} crafts.
We find that the effectiveness of the attack 
is not due to the 
amount of perturbations made on the input text, 
but rather how perturbations are made.
Specifically, we find two critical characteristics 
present in a vast majority of successful samples: 
(1) subject-predicate disagreement, 
meaning that a subject and corresponding verb within a sentence do not match, 
and (2) the changing of named entities. 
%
These characteristics 
are highlighted in~\cite{Interpret:Bertology}, 
where it was shown that BERT takes both 
into account when making predictions.
%

\emph{Fifth}, we test the effectiveness of potential countermeasures against adversarial slowdown.
We find that adversarial training~\citep{AT:NLP, A2T} is ineffective against \attname{}.
The 
defended multi-exit models lose efficacy,
or they lose significant amounts of accuracy 
in exchange for aggressive computational savings.
In contrast, we show that input sanitization
can be an effective countermeasure.
This result suggests that future work is needed 
to develop robust yet effective multi-exit language models.

%% file: sections/related.tex
\section{Related Work}
\label{sec:related-work}

\topic{Adversarial text attacks on language models.}
\citet{InitialAML} showed that neural network predictions
can be \emph{fool}-ed by human-imperceptible input perturbations
and called such perturbed inputs \emph{adversarial examples}.
While earlier work in this direction
studied these adversarial attacks on computer vision models~\cite{CWAttack, PGD},
there has been a growing body of work
on searching for adversarial examples in language domains 
as a result of language models gaining more traction.
However, 
adversarial texts are much harder to craft
due to the discrete nature of natural language.
Attacks on images leverage perturbations derived from computing \emph{gradients},
as they are composed of pixel values forming a near-continuous space,
but applying them to texts where each word is in a discrete space is not straightforward.
As a result, diverse mechanisms for 
crafting natural adversarial texts~\cite{HotFlip, GeneticAdv, TextBugger, TextFooler, BAE20, GDBA:21, CLARE:21} 
have been proposed.
In this work, we show that an adversary can exploit 
the language model's sensitivity to input text perturbations 
to achieve a completely different attack objective, i.e., adversarial slowdown.
A standard countermeasure against the adversarial input perturbation
is \emph{adversarial training} that augments the training data 
with natural adversarial texts~\cite{CertifiedRobustness, FreeLB, SMART, ALUM, A2T}.
We also show that adversarial training and its adaptation to our slowdown attacks
are ineffective in mitigating the vulnerability
to adversarial slowdown.

\topic{Input-adaptive efficient neural network inference.}
%
Neural networks are, while accurate,
computationally demanding in their post-training operations.
\citet{SDNOverthink} showed that
\emph{overthinking} is one problem%
---these models use all their internal layers
for making predictions on every single input
even for the ``easy" samples 
where a few initial layers would be sufficient.
Prior work~\cite{BranchyNet, SDNOverthink, MSDNets}
proposed \emph{multi-exit architectures}
to mitigate the wasteful effect of overthinking.
They introduce multiple internal classifiers
(i.e., early exits) to a pre-trained model 
and 
fine-tune them on the training data 
to make correct predictions.
During the inference on an input,
these early-exits enable \emph{input-adaptive} inference,
i.e., a 
model 
stops running forward 
if the prediction confidence 
is sufficient at an exit.

Recent work~\citep{RTRJ:20, DeeBERT, PABEE, ELBERT:21, BERxIT:21, LeeBERT:21, PastFuture, PCEE:22} adapted the multi-exit architectures 
to 
language models, e.g., BERT~\cite{BERT},
to 
save their 
computations at inference.
DeeBERT~\cite{DeeBERT} and FastBERT~\cite{FastBERT} have been proposed,
both straightforward adaptations of multi-exit architectures to language models.
\citet{PABEE} proposed patience-based early exits (PABEE)
and showed that one could achieve efficiency, accuracy, 
and robustness against natural adversarial texts. 
%
\citet{PastFuture} presented PastFuture
that makes predictions from a global perspective,
considering past and future predictions from all the exits.
However, no prior work studied the robustness of 
the computational savings that 
these mechanisms offer 
to adversarial slowdown~\cite{DeepSloth}.
We design a slowdown attack
to \emph{audit} their robustness.
More comparisons of our work to the work done in computer vision domains are in Appendix~\ref{appendix:novelty}.


%% file: sections/methods.tex
%
\section{Our Auditing Method: \attname{} Attack}
\label{sec:method}




\subsection{Threat Model}
\label{subsec:threat-model}

We consider an adversary who aims to reduce
the computational savings 
provided by a \emph{victim} multi-exit language model.
To achieve this goal, 
the attacker performs perturbations to a natural 
test-time text input $x \in \mathcal{S}$.
The resulting adversarial text $x'$ 
needs more layers to process for making a prediction.
This attack potentially violates
the computational guarantees made by real-time systems
harnessing multi-exit language models.
For example, the attacker can increase 
the operational costs of the victim model 
or push the response time of such systems outside the expected range.

Just like language models
deployed in the real-world
that accept any user input,
we assume that the attacker
has the ability to 
query the victim model with perturbed inputs.
%
We focus on the word-level perturbations as
they are well studied 
and efficient to perform 
with word embeddings~\cite{WordEmbedding}.
But it is straightforward to extend our attack to
character-level or sentence-level attacks 
by incorporating the slowdown objective 
we design in Sec~\ref{subsec:our-objective}
into their adversarial example-crafting algorithms.

To assess the slowdown vulnerability,
we comprehensively 
consider both \emph{white-box} and \emph{black-box} settings.
The white-box adversary
knows all the details of the victim model,
such as the training data and the model parameters,
while the black-box attacker has limited knowledge of the victim model.

\subsection{The Slowdown Objective}     
\label{subsec:our-objective}

Most adversarial text-crafting algorithms 
iteratively apply perturbations to a text input
until the resulting text $x'$ achieves a pre-defined goal.
The goal here for the standard adversarial attacks
is the untargeted misclassification of $x'$,
i.e., $f_{\theta}(x') \neq y$.
%
%
Existing adversarial text attacks 
design an objective (or a score) function that quantifies
how the perturbation of a word (e.g., substitution or removal)
helps the perturbed sample achieve the goal.
At each iteration $t$, the attacker considers all feasible perturbations
and chooses one that minimizes the objective the most.

We design our score function to quantify 
how close a perturbed sample $x'$ is to causing the worst-case slowdown.
The worst-case we consider here is that
$x'$ bypasses all the early exits, 
and the victim model classifies $x'$ at the final layer.
We formulate our score function $s(x', f_{\theta})$ as follows:
\begin{align*}
    s(x', f_{\theta}) = \sum_{0 < i \leq K} \Big( 1 - \frac{1}{N-1} \mathcal{L} \big( F_i(x'), \hat{y} \big) \Big)
\end{align*}
Here, the score function takes $x'$ as the perturbed text 
and $f_{\theta}$ as the victim model.
It computes the loss $\mathcal{L}$ between 
the softmax output of an $i$-th internal classifier $F_i$
and the uniform probability distribution $\hat{y}$ over classes.
We use $\ell_1$ loss.
$K$ is the number of early-exits,
and $N$ is the number of classes.

The score function $s$ 
returns a value in [0, 1].
%
It becomes one if all $F_i$ is close to the uniform distribution $\hat{y}$;
otherwise, it will be zero.
Unlike conventional adversarial attacks,
our score function over iterations
pushes all $F_i$ to $\hat{y}$
(\textit{i.e.}, the lowest confidence case).
Most early-exit mechanisms stop forward pass
if $F_i$'s prediction confidence is higher than a pre-defined threshold $T$;
thus, $x'$ that achieves the lowest confidence bypasses all the exit points.

\subsection{The \attname{} Attack}
\label{subsec:our-attack}

We finally implement \attname{}
by incorporating the slowdown objective we design 
into existing adversarial text attacks.
In this work, we adapt two existing attacks:
TextFooler~\cite{TextFooler} and A2T~\cite{A2T}.
TextFooler is by far the strongest black-box attack~\cite{AdvGLUE},
and A2T is a gradient-based white-box attack.
In particular, A2T can craft natural adversarial texts 
much faster than black-box attacks; 
thus, it facilitates adversarial training of language models.
We discuss the effectiveness of this in Sec~\ref{sec:defenses}.

We describe how we adapt TextFooler 
for auditing the slowdown risk in Algorithm~\ref{alg:textfooler}
(see Appendix for our adaptation of A2T).
We highlighted our adaptation to the original Textfooler in {\color{blue}blue}.

\input{algorithms/textfooler}

\topic{(line 1--6) Compute word importance.}
We first compute the importance of each word $w_i$ in a text input $x$.
TextFooler removes each word from $x$
and computes their influence on the final prediction result.
It then ranks the words based on their influence.
In contrast, we rank the words 
based on the \emph{increase} in the slowdown objective after each removal.
By perturbing only a few words,
we can minimize the alterations to $x$
and keep the semantic similarity between $x'$ and $x$.
Following the original study,
we filter out stop words, \textit{e.g.}, `the' or `when',
to minimize sentence structure destruction.

\topic{(line 7--9) Choose the set of candidate words for substitution.}
The attack then works by replacing a set of words
in $x$ with the candidates carefully chosen from $V$.
For each word $w_i \in x$,
the attack collects the set of $C$ candidates
by computing the top $N$ synonyms from $V$ (line 8).
It computes the cosine similarity 
between the embeddings of the word $w_i$ 
and the word $w' \in V$.
We use the same embeddings~\cite{WordEmbedding} 
that the original study uses.
TextFooler only keeps the candidate words
with the same part-of-speech (POS) as $w_i$
to minimize grammar destruction (line 9).

\topic{(line 10--28) Craft a natural slowdown text.}
We then iterate over the remaining candidates
and substitute $w_i$ with $c_k$.
If the text after this substitution $x^{temp}$ 
is sufficiently similar to the text before it, $x'$,
we store the candidate $c_k$ into $C_{final}$
and compute the slowdown score $s_k$.
In the end, we have a perturbed text input $x'$
that is similar to the original input within the $\epsilon$ similarity
and the slowdown score $s_k$ (line 10--17).
To compute the semantic similarity, 
we use Universal Sentence Encoder 
that the original study uses~\cite{USE}.

In line 20--26, if there exists any candidate $c_k$
that already increases the slowdown score $s_k$ over the threshold $\alpha$
we choose the word with the highest semantic similarity among these winning candidates.
However, when there is no such candidate,
we pick the candidate with the highest slowdown score,
substitute the candidate with $w_i$,
and repeat the same procedure with the next word $w_{i+1}$.
At the end (line 28),
TextFooler does not return any adversarial example if it fails to flip the prediction.
However, as our goal is causing slowdown, 
we use this adversarial text 
even when the score is $s_k \leq \alpha$.

%% file: algorithms/textfooler.tex
\begin{wrapfigure}{L}{0.58\linewidth}
\begin{minipage}{\linewidth}

\vspace{-2.2em} 

\begin{algorithm}[H]
\caption{\attname{} (based on TextFooler)}
\label{alg:textfooler}
\textbf{Input:} a text input $x = \{w_1, w_2, ..., w_n\}$, its label $y$, the victim model $f_{\theta}$, {\color{blue}its early exits $F_i$}, sentence similarity function $Sim(\cdot)$, its threshold $\epsilon$, word embeddings $E$ over the vocabulary $V$, and {\color{blue}the attack success threshold $\alpha$}. \\
\textbf{Output:} a natural adversarial text $x'$
\begin{algorithmic}[1]
    \State $x' \gets x$
    \For{each word $w_i$ in $x$}
        \State {\color{blue}Compute the importance $I_{w_i}$} 
    \EndFor
    
    \State Compose a set $W$ of all words $w_i \in x$ sorted by the descending order of their importance
    \State Remove the stop words from the set $W$
    
    \For{each word $w_i$ in $W$}
        \State \multiline{Initiate the set of substitute candidates $C$ by computing the top $N$ synonyms; we compute the cosine similarity between $E_{w_i}$ and $E_{w'}$, where $w' \in V$}
        \State $C \gets $\text{ POSFiler}$(C)$
        \State $C_{final} \gets \{\}$
        
        \For{$c_k$ in $C$}
            \State $x^{temp} \gets $\text{ Replace $w_j$ with $c_k$ in $x'$}
            \If{Sim$(x^{temp}, x') > \epsilon$}
                \State Add $c_k$ to $C_{final}$
                \State {\color{blue}$s_k \gets f_{\theta}(x^{temp})$}
            \EndIf
        \EndFor
        
        \If{{\color{blue}$\exists c_k$ whose score is $s_k \geq \alpha$}}
            \State Keep the candidates $c_k \in C_{final}$            
            \State $c^* \gets \argmax_{c \in C_{final}}$Sim$(x, x^{temp}_{w_j \rightarrow c})$
            \State $x' \gets $\text{ Replace $w_j$ with $c^*$ in $x'$}
            \State \Return $x'$
            
        \ElsIf{{\color{blue}$s_k(x')> min$ $s_k$}}
            \State {\color{blue}$c^* \gets \argmax_{c_k \in C_{final}} s_k$}
            \State $x' \gets $\text{ Replace $w_j$ with $c^*$ in $x'$}
            
        \EndIf
        
    \EndFor
    \State {\color{blue}\Return $x'$}
\end{algorithmic}
\end{algorithm}

\vspace{-2.8em} 

\end{minipage}
\end{wrapfigure}

%% file: sections/results.tex
%
\section{Auditing the Robustness to Adversarial Slowdown}
\label{sec:main-result}


We now utilize our \attname{} attack as a vehicle
to evaluate the robustness of the computational savings 
provided by multi-exit language models. 
Our adaptations of two adversarial text attacks, TextFooler and A2T, 
represent the black-box and white-box settings, respectively.


\noindent \textbf{Tasks.}
We evaluate the multi-exit language models 
trained on seven classification tasks chosen from
the GLUE benchmark~\citep{GLUE}: RTE, MRPC, MNLI, QNLI, QQP, SST-2, and CoLA.

\noindent \textbf{Multi-exit mechanisms.}
We consider three early-exit mechanisms 
recently proposed for language models:
DeeBERT~\citep{DeeBERT}, PABEE~\citep{PABEE}, and Past-Future~\citep{PastFuture}.
In DeeBERT, we take the pre-trained BERT and fine-tune it on the GLUE tasks.
We use the pre-trained ALBERT~\cite{ALBERT} for PABEE and Past-Future.
To implement these mechanisms, 
we use the source code from the original studies.
%
We describe all the implementation details, 
e.g., the hyper-parameter choices, 
in Appendix.

\noindent \textbf{Metrics.}
We employ two metrics: 
\emph{classification accuracy} and \emph{efficacy} 
proposed by~\citet{DeepSloth}.
We compute both the metrics 
on the test-set $S$ or the adversarial texts crafted on $S$.
Efficacy is a standardized metric
that quantifies a model's ability to use its early exits.
It is close to one
when a higher percentage of inputs exit at an early layer;
otherwise, it is 0.
To quantify the robustness,
we report the changes in accuracy and efficacy
of a clean test-set $S$ and $S$ perturbed using {\attname}.
%

\subsection{Multi-exit Language Models Are Not Robust to Adversarial Slowdown}
\label{subsec:major-findings}

\input{tables/slowdown_main}

Table~\ref{tab:main-results} shows our evaluation results.
Following the original studies,
we set the early-exit threshold, i.e., entropy or patience,
so that multi-exit language models have 0.33--0.5 efficacy
on the clean test set (see Appendix for more details).
We use four adversarial attacks:
two standard adversarial attacks, TextFooler (TF) and A2T,
and their 
adaptations: \attname{} (TF) and \attname{} (A2T).
%
We perform these attacks on the entire test-set
and report the changes in accuracy and efficacy.
In each cell, we include their flat changes
and the values computed on the clean and adversarial data in parenthesis.

\topic{Standard adversarial attacks are ineffective in auditing slowdown.}
We 
observe that the standard attacks (TF and A2T)
are ineffective in causing a significant slowdown.
In DeeBERT, those attacks cause negligible changes in efficacy (-0.05--0.14),
while they inflict a large accuracy drop (7\%--25\%).
Against PABEE and PastFuture,
we find that the changes are slightly higher than those observed from DeeBERT
(\textit{i.e.}, 0.02--0.13 and 0.03--0.31).
We can observe slowdowns in PastFuture,
but this is not because the standard attacks are effective in causing slowdowns.
This indicates the mechanism 
is more sensitive to input changes, 
which may lead to greater vulnerability to adversarial slowdown.

\topic{\attname{} is an effective auditing tool for assessing the slowdown risk.}
We show that our slowdown attack can inflict significant changes in efficacy.
In DeeBERT and PABEE, the attacks reduce the efficacy 0.06--0.26 and 0.07--0.29, respectively.
In PastFuture, we observe more reduction in efficacy 0.13--0.45.
These multi-exit language models are designed to achieve an efficacy of 0.33--0.5;
thus its reduction up to 0.29--0.45 means a complete offset of their computational savings.

\topic{The more complex a mechanism is, the more vulnerable it is to adversarial slowdown.}
\attname{} causes the most significant slowdown on PastFuture,
followed by PABEE and DeeBERT.
PastFuture stops forwarding
based on the predictions from past exits
and the estimated predictions from future exits.
PABEE also uses patience, 
i.e., how often we observe the same decision over early-exits.
They enable more \emph{aggressive} efficacy compared to DeeBERT, 
which only uses entropy.
However, this aggressiveness can be exploited by our attacks,
e.g., introducing inconsistencies over exit points;
thus, PABEE needs more layers to make a prediction.

\subsection{Sensitivity to Attack Hyperparameter}
\label{subsec:ablation-study}

\begin{wrapfigure}{r}{0.5\linewidth}
    \vspace{-2.4em}
    \centering
    \includegraphics[width=\linewidth]{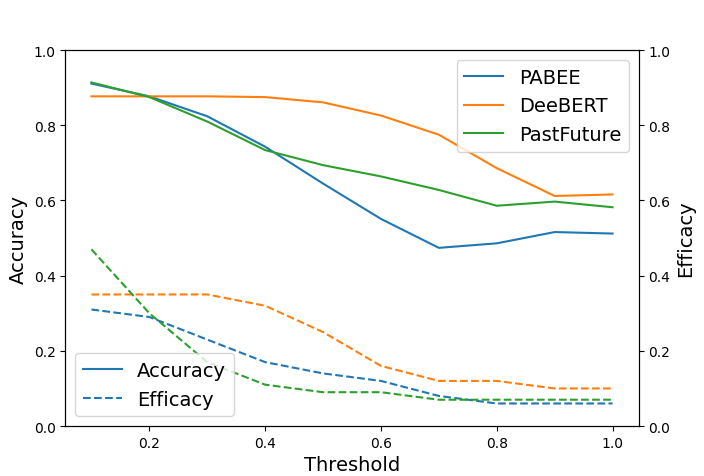}
    \caption{\textbf{The impact of $\alpha$ on accuracy and efficacy.} Taking each model's results 
    on QNLI, 
    as $\alpha$ is increased, the accuracy and efficacy 
    decrease.}
    \label{fig:threshold}
    \vspace{-2.em}     
\end{wrapfigure}

The key hyperparameter of our attack, 
the attack success threshold ($\alpha$),
determines the magnitude of the scores 
pursued by \attname{} while crafting adversarials.
The score measures
how uniform all output distributions of $F_i$ are.
A higher $\alpha$ pushes \attname{} 
to have a higher slowdown score 
before returning a perturbed sample.
Figure 2 shows 
the accuracy and efficacy of all three mechanisms on QNLI
against $\alpha$ in [0.1, 1].
We show that as $\alpha$ increases, 
the slowdown (represented as a decrease in efficacy) increases, 
and the accuracy decreases.

In addition, as $\alpha$ increases, 
the rate of decrease in accuracy and efficacy decreases.
Note that in PastFuture,
when $\alpha\!\geq\!0.4$ the rate at which efficacy decreases drops by a large margin. The same applies to accuracy, and when $\alpha\!\geq\!0.8$, accuracy surprisingly increases, a potentially undesirable outcome.
%
%
Moreover, 
when $\alpha\!\geq\!0.8$ efficacy does not decrease any further, 
which potentially wastes computational resources as the time required to craft samples increases greatly as $\alpha$ is increased.
%

%
\input{sections/blackbox}

%
\section{Lingusitic Analysis of Our Adversarial Texts}
\label{sec:qualitative-study}

\begin{wrapfigure}{r}{0.44\linewidth}
    \vspace{-1.2em}     
    \centering
    \includegraphics[width=\linewidth]{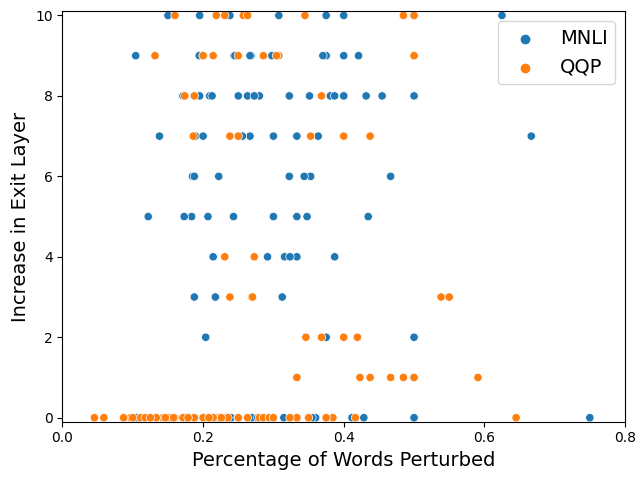}
    \vspace{-1.5em}     
    \caption{\textbf{Visualizing the relationship between words perturbed and adversarial slowdown induced.} 
    Taking 100 random samples from MNLI and QQP (crafted against the PastFuture model), 
    the percentage of words that \attname{} perturbed and the resulting increase in exit layer is plotted (a greater increase in exit layer indicates greater slowdown). 
    The figure shows that the percentage of words perturbed has \emph{no} influence over a sample's ability to induce slowdown.}
    \label{fig:perturbation-n-slowdown}
    \vspace{-2.em}     
\end{wrapfigure}

To qualitatively analyze the text generated by \attname{}, 
we first consider how the number of perturbations applied to an input text affects the slowdown it induces. 
We 
choose samples crafted against PastFuture \cite{PastFuture}, 
due to it being the most vulnerable to our attack. 
We 
select the datasets that induce the most and least slowdown, MNLI and QQP, respectively, 
in order to conduct a well-rounded analysis.
Using 100 samples randomly drawn from both datasets, 
we record the percentage of words perturbed by \attname{} 
and the consequent increase in exit layer.
In Figure~\ref{fig:perturbation-n-slowdown}, 
we 
find that there is no relationship between the percentage of words perturbed and the increase in exit layer.
It is not the number of perturbations made that affects slowdown, but rather how perturbations are made.

In an effort to 
find another explanation for why samples crafted by \attname{} induce slowdown on multi-exit models, 
we look to analyze the inner workings of BERT.
Through the qualitative analysis performed by \citet{Interpret:Bertology}, 
we find particular interest in two characteristics deemed of high importance to BERT 
when it makes predictions: (1) subject-predicate agreement and (2) the changing of named entities.
In 
our experiment, 
these characteristics are incredibly prevalent in successful 
attacks.
Particularly, we find that the score assigned by \attname{} is much higher 
when the subject and predicate of a sentence do not match, or a named entity is changed.
In addition, \attname{} often changes non-verbs into verbs or removes verbs entirely, 
causing further subject-predicate discrepancies.
Table~\ref{tab:examples} shows an example of these characteristics across all the GLUE tasks. 
Note that subject-predicate agreement appears much more often. 

\input{tables/slowdown_examples}

To quantify the prevalence of these characteristics,
we analyze 100 adversarial texts, from QQP crafted on DeeBert,
most effective in inducing slowdowns
and count the number of samples containing subject-predicate disagreement 
or a changed named entity.
Of the 100 samples, 84\% had a subject-predicate disagreement 
and 31\% a changed named entity.
An important note is that 
we see smaller percentage points in a changed named entity 
than in a subject-predicate disagreement, 
as not all samples have a named entity.

We believe that these two characteristics induce a high amount of slowdown 
because they can 
push samples toward out-of-distribution.
BERT was trained to have an acute understanding of language and its nuances.
It is therefore not expected to have the disagreement 
between the subject and predicate of a sentence in its training data
(unless that is the task it is trained for e.g. CoLA).
Also, according to~\citet{Interpret:Bertology}, 
BERT likely lacks a general idea of named entities, 
providing an explanation as to why the model is less confident in an answer 
when a named entity is changed.


Moreover, we analyze samples produced by the base attacks (TF and A2T), 
and compare them with samples produced by \attname{}.
%
Surprisingly, we find a great deal of similarity, 
particularly with regard to the usage of the two characteristics detailed above, 
despite the base attacks inducing negligible slowdown.
A potential explanation for this is that 
both types of attacks construct out-of-distribution samples,
but change a model's confidence to varying extents.
The base attacks likely increase confidence in the wrong answer 
to a much larger degree than \attname{}, 
encouraging early-exits with mispredictions in a majority of cases.
In contrast, \attname{} 
works by reducing confidence across all of the classes.

%% file: tables/slowdown_main.tex
%
%
\begin{table*}[ht]
\adjustbox{max width=\textwidth}{%
    \begin{tabular}{@{}r|c|ccccccc@{}}
        \toprule
        \multicolumn{1}{c|}{\multirow{2}{*}{\textbf{Attack}}} 
         & \multirow{2}{*}{\textbf{Metric}} 
         & \multicolumn{7}{c}{\textbf{GLUE Task}} \\ \cmidrule(l){3-9} 
         &  
         & \multicolumn{1}{c}{\textbf{RTE}} 
         & \multicolumn{1}{c}{\textbf{MNLI}} 
         & \multicolumn{1}{c}{\textbf{MRPC}} 
         & \multicolumn{1}{c}{\textbf{QNLI}} 
         & \multicolumn{1}{c}{\textbf{QQP}} 
         & \multicolumn{1}{c}{\textbf{SST-2}} 
         & \multicolumn{1}{c}{\textbf{CoLA}} \\ \midrule \midrule
        \multicolumn{9}{c}{\textbf{DeeBERT (BERT-base)}} \\ \midrule
        \multirow{2}{*}{\textbf{TF}} 
            & \textbf{Acc.}
            & 63 $\rightarrow$ 48 & \multicolumn{1}{c}{-}
            & 82 $\rightarrow$ 75 & 88 $\rightarrow$ 78 
            & 92 $\rightarrow$ 67 & \multicolumn{1}{c}{-}
            & 79 $\rightarrow$ 57 \\ 
            & \textbf{Eff.}
            & 0.34 $\rightarrow$ 0.32 & \multicolumn{1}{c}{-}
            & 0.35 $\rightarrow$ 0.32 & 0.35 $\rightarrow$ 0.33 
            & 0.36 $\rightarrow$ 0.40 & \multicolumn{1}{c}{-}
            & 0.34 $\rightarrow$ 0.20 \\ \midrule
        \multirow{2}{*}{\textbf{A2T}} 
            & \textbf{Acc.}
            & 63 $\rightarrow$ 52 & \multicolumn{1}{c}{-}
            & 82 $\rightarrow$ 75 & 88 $\rightarrow $81
            & 92 $\rightarrow$ 74 & \multicolumn{1}{c}{-}
            & 79 $\rightarrow$ 66 \\
            & \textbf{Eff.}
            & 0.34 $\rightarrow$ 0.32 & \multicolumn{1}{c}{-}
            & 0.35 $\rightarrow$ 0.33 & 0.35 $\rightarrow$ 0.35
            & 0.36 $\rightarrow$ 0.41 & \multicolumn{1}{c}{-}
            & 0.34 $\rightarrow$ 0.29 \\ \midrule
        \multirow{2}{*}{\textbf{\attname{} (TF)}} 
            & \textbf{Acc.}
            & 63 $\rightarrow$ 51 & \multicolumn{1}{c}{-}
            & 82 $\rightarrow$ 61 & 88 $\rightarrow$ 62
            & 92 $\rightarrow$ 69 & \multicolumn{1}{c}{-}
            & 79 $\rightarrow$ 70 \\
            & \textbf{Eff.}
            & 0.34 $\rightarrow$ 0.11 & \multicolumn{1}{c}{-}
            & 0.35 $\rightarrow$ 0.09 & 0.35 $\rightarrow$ 0.10 
            & 0.36 $\rightarrow$ 0.22 & \multicolumn{1}{c}{-}
            & 0.34 $\rightarrow$ 0.13 \\ \midrule
        \multirow{2}{*}{\textbf{\attname{} (A2T)}} 
            & \textbf{Acc.}
            & 63 $\rightarrow$ 57 & \multicolumn{1}{c}{-}
            & 82 $\rightarrow$ 75 & 88 $\rightarrow$ 78
            & 92 $\rightarrow$ 83 & \multicolumn{1}{c}{-}
            & 79 $\rightarrow$ 73 \\
            & \textbf{Eff.}
            & 0.34 $\rightarrow$ 0.19 & \multicolumn{1}{c}{-}
            & 0.35 $\rightarrow$ 0.17 & 0.35 $\rightarrow$ 0.19 
            & 0.36 $\rightarrow$ 0.30 & \multicolumn{1}{c}{-}
            & 0.34 $\rightarrow$ 0.24 \\ \midrule \midrule
        \multicolumn{9}{c}{\textbf{PABEE (ALBERT-base)}} \\ \midrule
        \multirow{2}{*}{\textbf{TF}} 
            & \textbf{Acc.}
            & 79 $\rightarrow$ 34 & 83 $\rightarrow$ 25
            & 87 $\rightarrow$ 37 & 91 $\rightarrow$ 33
            & 92 $\rightarrow$ 31 & 93 $\rightarrow$ 22
            & 82 $\rightarrow$ 5 \\
            & \textbf{Eff.}
            &  0.24 $\rightarrow$ 0.22 & 0.28 $\rightarrow$ 0.17 
            &  0.32 $\rightarrow$ 0.21 & 0.31 $\rightarrow$ 0.18
            &  0.37 $\rightarrow$ 0.27 & 0.37 $\rightarrow$ 0.26
            &  0.32 $\rightarrow$ 0.23 \\ \midrule
        \multirow{2}{*}{\textbf{A2T}} 
            & \textbf{Acc.}
            & 79 $\rightarrow$ 57 & 83 $\rightarrow$ 52
            & 87 $\rightarrow$ 63 & 91 $\rightarrow$ 71
            & 92 $\rightarrow$ 61 & 93 $\rightarrow$ 76
            & 82 $\rightarrow$ 38 \\
            & \textbf{Eff.}
            &  0.24$ \rightarrow$ 0.22 & 0.28$ \rightarrow$ 0.21 
            &  0.32$ \rightarrow$ 0.26 & 0.31$ \rightarrow$ 0.27
            &  0.37$ \rightarrow$ 0.31 & 0.37$ \rightarrow$ 0.32
            &  0.32$ \rightarrow$ 0.23 \\ \midrule
        \multirow{2}{*}{\textbf{\attname{} (TF)}} 
            & \textbf{Acc.}
            & 79 $\rightarrow$ 57 & 83 $\rightarrow$ 38
            & 87 $\rightarrow$ 47 & 91 $\rightarrow$ 51
            & 92 $\rightarrow$ 67 & 93 $\rightarrow$ 50
            & 82 $\rightarrow$ 48 \\
            & \textbf{Eff.}
            &  0.24 $\rightarrow$ 0.09 & 0.28 $\rightarrow$ 0.05 
            &  0.32 $\rightarrow$ 0.08 & 0.31 $\rightarrow$ 0.06
            &  0.37 $\rightarrow$ 0.17 & 0.37 $\rightarrow$ 0.08
            &  0.32 $\rightarrow$ 0.08 \\ \midrule
        \multirow{2}{*}{\textbf{\attname{} (A2T)}} 
            & \textbf{Acc.}
            & 79 $\rightarrow$ 72 & 83 $\rightarrow$ 69
            & 87 $\rightarrow$ 73 & 91 $\rightarrow$ 82
            & 92 $\rightarrow$ 79 & 93 $\rightarrow$ 85
            & 82 $\rightarrow$ 60 \\
            & \textbf{Eff.}
            &  0.24 $\rightarrow$ 0.17 & 0.28 $\rightarrow$ 0.18 
            &  0.32 $\rightarrow$ 0.21 & 0.32 $\rightarrow$ 0.23  
            &  0.37 $\rightarrow$ 0.27 & 0.37 $\rightarrow$ 0.29
            &  0.32 $\rightarrow$ 0.19 \\ \midrule \midrule
        \multicolumn{9}{c}{\textbf{PastFuture (ALBERT-base)}} \\ \midrule
        \multirow{2}{*}{\textbf{TF}} 
            & \textbf{Acc.}
            & 74 $\rightarrow$ 41 & 86 $\rightarrow$ 42
            & 88 $\rightarrow$ 36 & 92 $\rightarrow$ 52
            & 92 $\rightarrow$ 50 & \multicolumn{1}{c}{-}
            & \multicolumn{1}{c}{-} \\
            & \textbf{Eff.}
            &  0.52 $\rightarrow$ 0.46 & 0.50 $\rightarrow$ 0.24
            &  0.50 $\rightarrow$ 0.24 & 0.50 $\rightarrow$ 0.19
            &  0.52 $\rightarrow$ 0.35 & \multicolumn{1}{c}{-}
            & \multicolumn{1}{c}{-} \\ \midrule
        \multirow{2}{*}{\textbf{A2T}} 
            & \textbf{Acc.}
            & 74 $\rightarrow$ 58 & 86 $\rightarrow$ 59
            & 88 $\rightarrow$ 58 & 92 $\rightarrow$ 74
            & 92 $\rightarrow$ 64 & \multicolumn{1}{c}{-}
            & \multicolumn{1}{c}{-} \\
            & \textbf{Eff.}
            &  0.52 $\rightarrow$ 0.49 & 0.50 $\rightarrow$ 0.32
            &  0.50 $\rightarrow$ 0.31 & 0.50 $\rightarrow$ 0.35
            &  0.52 $\rightarrow$ 0.43 & \multicolumn{1}{c}{-}
            & \multicolumn{1}{c}{-} \\ \midrule
        \multirow{2}{*}{\textbf{\attname{} (TF)}} 
            & \textbf{Acc.}
            & 74 $\rightarrow$ 51 & 86 $\rightarrow$ 45
            & 88 $\rightarrow$ 42 & 92 $\rightarrow$ 58
            & 92 $\rightarrow$ 64 & \multicolumn{1}{c}{-}
            & \multicolumn{1}{c}{-} \\
            & \textbf{Eff.}
            &  0.51 $\rightarrow$ 0.17 & 0.50 $\rightarrow$ 0.05
            &  0.50 $\rightarrow$ 0.15 & 0.50 $\rightarrow$ 0.07
            &  0.52 $\rightarrow$ 0.25 & \multicolumn{1}{c}{-}
            & \multicolumn{1}{c}{-} \\ \midrule
        \multirow{2}{*}{\textbf{\attname{} (A2T)}} 
            & \textbf{Acc.}
            & 74 $\rightarrow$ 64 & 86 $\rightarrow$ 67
            & 88 $\rightarrow$ 72 & 92 $\rightarrow$ 83
            & 92 $\rightarrow$ 79 & \multicolumn{1}{c}{-}
            & \multicolumn{1}{c}{-} \\
            & \textbf{Eff.}
            &  0.52 $\rightarrow$ 0.36 & 0.50 $\rightarrow$ 0.26
            &  0.50 $\rightarrow$ 0.29 & 0.50 $\rightarrow$ 0.33
            &  0.52 $\rightarrow$ 0.39 & \multicolumn{1}{c}{-}
            & \multicolumn{1}{c}{-} \\ \bottomrule
    \end{tabular}
}
\caption{%
    \textbf{Robustness of multi-exit language models to our slowdown attacks.} 
    \attname{} significantly reduces the computational savings offered by DeeBERT, PABEE, and PastFuture.
    In each cell, we report the accuracy (acc.) and efficacy (eff.) on a clean test set and its corresponding adversarial texts generated by the four attacks ($\rightarrow$ denotes going from the clean test set to the adversarial texts).
}
\label{tab:main-results}
%
%
\end{table*}

%% file: sections/blackbox.tex
\section{Practical Exploitation of \attname{} in Black-box Settings}
\label{sec:black-box-attacks}

In Sec~\ref{sec:main-result}, 
we show in the worst-case scenarios, 
multi-exit language models are not robust to adversarial slowdown.
We now turn our attention to black-box settings 
where an adversary does not have full knowledge of the victim's system.
We consider two attack scenarios: 
(1) \emph{Transfer-based attacks}
where an adversary who has the knowledge of the training data 
trains \emph{surrogate} models to craft adversarial texts 
and use them to attack the victim models.
(2) \emph{Universal attacks} 
where an attacker finds a set of \emph{trigger} words 
that can inflict slowdown when attached to any test-time inputs. 
We run these experiments across various GLUE tasks
and show the results from the RTE and QQP datasets.
We include all our results 
on different datasets and victim-surrogate combinations 
in the Appendix.


\topic{Transferability of \attname{}.}
We first test if our attack is transferable
in three different scenarios:
(1) Cross-seed; (2) Cross-architecture; and (3) Cross-mechanism.
Table~\ref{tbl:black-box} summarizes our results.



\begin{table*}[ht]
\centering
\begin{threeparttable}
\adjustbox{max width=\textwidth}{
    \begin{tabular}{@{}c|c|c|c||c|cc|cc@{}}
    \toprule
    \multirow{2}{*}{\textbf{Model}} & \multirow{2}{*}{\textbf{Arch.}} & \multirow{2}{*}{\textbf{Mechanism}} & \multirow{2}{*}{\textbf{Scenario}} & \multirow{2}{*}{\textbf{Type}} & \multicolumn{2}{c|}{\textbf{RTE}} & \multicolumn{2}{c}{\textbf{QQP}} \\ 
    &  &  &  &  & \textbf{Acc.} & \textbf{Eff.} & \textbf{Acc.} & \textbf{Eff.} \\ \midrule \midrule
    \textbf{S} & \textbf{BERT} & \textbf{PastFuture} & \multirow{2}{*}{\textbf{Cross-seed}} & S$\rightarrow$S & 66 $\rightarrow$ 52 & 0.47 $\rightarrow$ 0.11 & 91 $\rightarrow$ 72 & 0.50 $\rightarrow$ 0.26 \\
    \textbf{V} & \textbf{BERT} & \textbf{PastFuture} & & S$\rightarrow$V & 66 $\rightarrow$ 55 & 0.50 $\rightarrow$ 0.25 & 91 $\rightarrow$ 72 & 0.51 $\rightarrow$ 0.33 \\ \midrule
    \textbf{S} & \textbf{BERT} & \textbf{PABEE} & \multirow{2}{*}{\textbf{Cross-arch.}} & S$\rightarrow$S & 66 $\rightarrow$ 49 & 0.22 $\rightarrow$ 0.08 & 91 $\rightarrow$ 69 & 0.35 $\rightarrow$ 0.16 \\
    \textbf{V} & \textbf{ALBERT} & \textbf{PABEE} &  & S$\rightarrow$V & 77 $\rightarrow$ 55 & 0.22 $\rightarrow$ 0.21 & 91 $\rightarrow$ 74 & 0.36 $\rightarrow$ 0.34 \\ \midrule
    \textbf{S} & \textbf{BERT} & \textbf{PABEE} & \multirow{2}{*}{\textbf{Cross-mech.}} & S$\rightarrow$S & 66 $\rightarrow$ 49 & 0.22 $\rightarrow$ 0.08 & 91 $\rightarrow$ 69 & 0.35 $\rightarrow$ 0.16 \\
    \textbf{V} & \textbf{BERT} & \textbf{PastFuture} &  & S$\rightarrow$V & 66 $\rightarrow$ 55 & 0.50 $\rightarrow$ 0.29 & 91 $\rightarrow$ 74 & 0.51 $\rightarrow$ 0.38 \\ \bottomrule
    \end{tabular}
}
\begin{tablenotes}
    \item S = Surrogate model; V = Victim model
\end{tablenotes}
\end{threeparttable}
\caption{\textbf{Transfer-based attack results.} Results from the cross-seed, cross-mechanism, and cross-architecture experiments on RTE and QQP. In all experiments, we craft adversarial texts on the surrogate model (S) and then evaluated on both the surrogate (S$\rightarrow$S) and victim (S$\rightarrow$V) models.}
\label{tbl:black-box}
\end{table*}

\emph{Cross-seed.} 
Both the model architecture and early-exit mechanism are identical 
for the surrogate and the victim models. 
In RTE and QQP, our transfer attack (S$\rightarrow$V) demonstrates
a significant slowdown on the victim model, 
resulting in a reduction in efficacy of 0.25 and 0.18, respectively. 
In comparison to the white-box scenarios (S$\rightarrow$S), 
these attacks achieve approximately 50\% effectiveness.

\emph{Cross-architecture.}
We vary the model architecture, using either BERT or ALBERT, 
while keeping the early-exit mechanism (PABEE) the same.
Across the board, we achieve the lowest transferability among the three attacking scenarios, 
with a reduction in efficacy of 0.01 in RTE and 0.02 in QQP, respectively. 
This indicates that when conducting transfer-based attacks, 
the matching of the victim and surrogate models' architectures 
has a greater impact than the early-exit mechanism.

\emph{Cross-mechanism.}
We now vary the early-exit mechanism used by the victim and surrogate models 
while the architecture (BERT) remains consistent. 
In QQP and RTE, we cause significant slowdown to the victim model 
(a reduction in efficacy of 0.21 and 0.13, respectively),
even when considering the relational speed-up offered by different mechanisms 
(e.g., PastFuture offers more computational savings than PABEE and DeeBERT).
The slowdown is comparable to the white-box cases (S$\rightarrow$S).

%

\topic{Universal slowdown triggers.}
If the attacker is unable to train surrogate models,
they can find a few words (i.e., a trigger) that causes slowdown 
to any test-time inputs when attached.
Searching for such a trigger does not require the knowledge of the training data.
%
%
To demonstrate the practicality of this attack, 
we select 1000 random words from BERT's vocabulary 
and compute the total slowdown across 10\% of the SST-2 test dataset 
by appending each vocab word to the beginning of every sentence. 
We then choose the word that induces the highest slowdown and evaluate it against the entire test dataset.
We find that the most effective word, "unable", reduces efficacy by 9\% and accuracy by 14\% 
when appended to the beginning of all sentences once.
When appended three times successively (i.e. "unable unable unable \ldots"), 
the trigger reduces efficacy by 18\% and accuracy by 3\%.


%% file: tables/slowdown_examples.tex
\begin{table*}[t]
\adjustbox{max width=\textwidth} {
\begin{tabular}{@{}c|l|l|c|c@{}}
\toprule
\textbf{Task} &
  \multicolumn{1}{c|}{\textbf{Original Text}} &
  \multicolumn{1}{c|}{\textbf{Perturbed Text}} &
  \multicolumn{2}{c}{\textbf{Change}} \\ \midrule \midrule
\textbf{RTE} &
  \begin{tabular}[c]{@{}l@{}}Sentence1: Mount Olympus towers up from the center of the earth.\\ Sentence2: Mount Olympus is in the center of the earth.\end{tabular} &
  \begin{tabular}[c]{@{}l@{}}Sentence1: {\color{red}\textbf{Install}} Olympus towers up from the {\color{red}\textbf{facilities}} of the {\color{red}\textbf{planet}}.\\ Sentence2: Mount Olympus is in the center of the earth.\end{tabular} &
  \multicolumn{2}{c}{8$\rightarrow$12} 
  \\ \midrule
\textbf{MNLI} &
  \begin{tabular}[c]{@{}l@{}}Premise: I'll twist him, sir. \\ Hypothesis: I'll make him straight\end{tabular} &
  \begin{tabular}[c]{@{}l@{}}Premise: I'll {\color{red}\textbf{bending}} him, sir.\\ Hypothesis: I'll {\color{red}\textbf{implement}} him {\color{red}\textbf{consecutive}}.\end{tabular} &
  \multicolumn{2}{c}{8$\rightarrow$12} 
  \\ \midrule
\textbf{MRPC} & 
  \begin{tabular}[c]{@{}l@{}}Sentence1: Ms Stewart, the chief executive, was not expected to attend.\\ Sentence2: Ms Stewart, 61, ..., did not attend.\end{tabular} &
  \begin{tabular}[c]{@{}l@{}}Sentence1: {\color{red}\textbf{Lena}} Stewart, the chief {\color{red}\textbf{execute}}, was not {\color{red}\textbf{scheduled}} to {\color{red}\textbf{help}}.\\ Sentence2: Ms Stewart, 61, ..., did not attend.\end{tabular} &
  \multicolumn{2}{c}{7$\rightarrow$12} 
  \\ \midrule
\textbf{QNLI} & 
  \begin{tabular}[c]{@{}l@{}}Question: Where did the Exposition take place?\\ Sentence: This World's Fair devoted a building to electrical exhibits.\end{tabular} &
  \begin{tabular}[c]{@{}l@{}}Question: {\color{red}\textbf{Whereby}} did the {\color{red}\textbf{Shows}} {\color{red}\textbf{takes}} place?\\ Sentence: This World's Fair {\color{red}\textbf{devoting}} a building to electrical exhibits.\end{tabular} &
  \multicolumn{2}{c}{7$\rightarrow$12} 
  \\ \midrule
\textbf{QQP} &
  \begin{tabular}[c]{@{}l@{}}Question1: Why do we need to philosophize?\\ Question2: Why do we need to philosophize with others?\end{tabular} &
  \begin{tabular}[c]{@{}l@{}}Question1: Why do we need to philosophize?\\ Question2: Why {\color{red}\textbf{got}} we {\color{red}\textbf{needing}} to philosophize with others?\end{tabular} &
  \multicolumn{2}{c}{10$\rightarrow$12} 
  \\ \midrule
\textbf{SST-2} &
  it's a cookie-cutter movie, a cut-and-paste job. &
  it's a cookie-cutter {\color{red}\textbf{cinematography}}, a cut-and-paste {\color{red}\textbf{worked}}. &
  \multicolumn{2}{c}{7$\rightarrow$12} 
  \\ \midrule
\textbf{CoLA} & 
  I'll work on it if I can. &
  I'll {\color{red}\textbf{task}} on it if {\color{red}\textbf{me}} can. &
  \multicolumn{2}{c}{7$\rightarrow$12} 
  \\ \bottomrule
\end{tabular}
}
\caption{%
    \textbf{Adversarial texts generated by \attname{}.} An example of text perturbed by \attname{} from each dataset, using samples crafted for PABEE. 
    They show how subject-predicate disagreement and the changing of named entities can push previously early-exited samples to the final output layer.
}
\label{tab:examples}
\end{table*}

%% file: sections/discussion.tex
\section{Potential Countermeasures}
\label{sec:defenses}


We 
now test the effectiveness of potential countermeasures against adversarial slowdown.
We first evaluate \emph{adversarial training} (AT), 
a standard countermeasure that reduces the sensitivity of a model 
to adversarial input perturbations.
We then discuss a way to \emph{sanitize} the perturbations applied to inputs
by running off-the-shelf tools, such as a grammar-checking tool, to remove attack artifacts.

To evaluate, we use the AT proposed by~\citet{A2T}.
This AT requires significantly fewer computational resources;
thus, it is better suited for adversarially training large language models.
We run AT with two different natural adversarial texts,
those crafted by A2T and by \attname{} (adapted from A2T).
%
During training, we attack 20\% of the total samples in each batch.
We run our experiments with PABEE trained on RTE and SST-2.
We first set the patience to 6 (consistent with the rest of our experiments),
but we set it to 2 for the models trained with \attname{} (A2T).
Once trained, we examine the defended models 
with attacks stronger than A2T: 
TextFooler (TF) and \attname{} (Ours).

\input{tables/slowdown_at_results}

\topic{AT is ineffective against our slowdown attacks.}
Table~\ref{tbl:at-results} shows 
that AT significantly reduces the efficacy of a model.
Compared to the undefended models,
the defended models achieve $\sim$0 efficacy.
As these models do not utilize early exits,
they seem robust to our attacks.
But certainly, it is not desirable.
It is noticeable that
the defended models still suffer from a large accuracy drop.
We then decided to set the patience to two,
i.e., the multi-exit language models use early-exits more aggressively.
The defended models have either 
very low efficacy or accuracy,
and our attacks can reduce both.
This result highlights a trade-off between
being robust against adversarial slowdown and being efficient.
We leave further explorations as future work.

\topic{Input sanitization can be a defense against adversarial slowdown.}
Our linguistic analysis in Sec~\ref{sec:qualitative-study} shows that 
the subject-predicate discrepancy is one of the root causes of the slowdown.
Building on this insight,
we test if sanitizing the perturbed input before feeding it into the models
can be a countermeasure against our slowdown attacks.
We evaluate this hypothesis with two approaches.

We first use OpenAI's ChatGPT\footnote{ChatGPT: \href{https://openai.com/blog/chatgpt/}{https://openai.com/blog/chatgpt/}},
a conversational model where we can ask questions and get answers.
We manually query the model with 
natural adversarial texts generated by \attname{} (TF)
and collect the revised texts.
Our query starts with the prompt 
``Can you fix all of these?" 
followed by perturbed texts in the subsequent lines.
We evaluate with the MNLI and QQP datasets,
on 50 perturbed test-set samples randomly chosen from each.
We compute the changes in accuracy and average exit number
on the perturbed samples and their sanitized versions.
We compute them on the PastFuture models trained on the datasets.
Surprisingly, we find that the inputs sanitized by ChatGPT 
greatly recovers both accuracy and efficacy.
In MNLI, we recover the accuracy by 12 percentage points (54\%$\rightarrow$66\%) 
and reduce the average exit number by 4 (11.5$\rightarrow$7.5).
In QQP, the accuracy is increased by 24 percentage points (56\%$\rightarrow$80\%), 
and the average exit number is reduced by 2.5 (9.6$\rightarrow$7.1).
%

We also test the effectiveness of additional grammar-checking tools, such as
Grammarly\footnote{Grammarly: \href{https://www.grammarly.com/}{https://www.grammarly.com/}} 
and language\_tool\_python\footnote{Language Tool (Python): \href{https://github.com/jxmorris12/language_tool_python}{https://github.com/jxmorris12/language\_tool\_python}},
in defeating our slowdown attacks.
We run this evaluation using the same settings as mentioned above.
We feed the adversarial texts generated by our attack into Grammarly and have it correct them.
Note that we only take the Grammarly recommendations for the correctness
and disregard any other recommendations, such as for clarity.
We find that the inputs sanitized by Grammarly
still suffer from a significant accuracy loss and slowdown.
In MNLI, both the accuracy and the average exit number 
stay the same 56\%$\rightarrow$58\% and 9.6$\rightarrow$9.5, respectively.
In QQP, we observe that there is
almost no change in accuracy (54\%$\rightarrow$58\%)
or the average exit number (11.5$\rightarrow$11.5).


%% file: tables/slowdown_at_results.tex
\begin{wraptable}{R}{0.6\linewidth}
\vspace{-1.em}     
\centering
\adjustbox{max width=\linewidth}{%
    \begin{tabular}{@{}c|c|c|cc|cc@{}}
        \toprule
        \multirow{2}{*}{\textbf{AT}} & \multirow{2}{*}{\textbf{P}} & \multirow{2}{*}{\textbf{Attack}} & \multicolumn{2}{c|}{\textbf{RTE}} & \multicolumn{2}{c}{\textbf{SST-2}} \\ \cmidrule(l){4-7} 
         &  &  & \multicolumn{1}{c}{\textbf{Acc.}} & \multicolumn{1}{c|}{\textbf{Eff.}} & \multicolumn{1}{c}{\textbf{Acc.}} & \multicolumn{1}{c}{\textbf{Eff.}} \\ \midrule \midrule
        \multirow{4}{*}{\textbf{A2T}} & \multirow{2}{*}{\textbf{6}} & \textbf{TF} & 81 $\rightarrow$ 8 & 0.04 $\rightarrow$ 0.04 & 92 $\rightarrow$ 5 & 0.04 $\rightarrow$ 0.04 \\
         &  & \textbf{Ours} & 81 $\rightarrow$ 60 & 0.04 $\rightarrow$ 0.04 & 92 $\rightarrow$ 59 & 0.04 $\rightarrow$ 0.04 \\ \cmidrule(l){2-7} 
         & \multirow{2}{*}{\textbf{2}} & \textbf{TF} & 72 $\rightarrow$ 24 & 0.13 $\rightarrow$ 0.13 & 89 $\rightarrow$ 10 & 0.08 $\rightarrow$ 0.07 \\
         &  & \textbf{Ours} & 72 $\rightarrow$ 59 & 0.13 $\rightarrow$ 0.14 & 89 $\rightarrow$ 56 & 0.08 $\rightarrow$ 0.07 \\ \midrule \midrule
        \multirow{4}{*}{\textbf{\begin{tabular}[c]{@{}c@{}}A2T\\ (Ours)\end{tabular}}} & \multirow{2}{*}{\textbf{6}} & \textbf{TF} & 78 $\rightarrow$ 7 & 0.04 $\rightarrow$ 0.04 & 92 $\rightarrow$ 6 & 0.04 $\rightarrow$ 0.04 \\
         &  & \textbf{Ours} & 78 $\rightarrow$ 56 & 0.04 $\rightarrow$ 0.04 & 92 $\rightarrow$ 61 & 0.04 $\rightarrow$ 0.04 \\ \cmidrule(l){2-7} 
         & \multirow{2}{*}{\textbf{2}} & \textbf{TF} & 53 $\rightarrow$ 53 & 0.65 $\rightarrow$ 0.65 & 90 $\rightarrow$ 7 & 0.05 $\rightarrow$ 0.04 \\
         &  & \textbf{Ours} & 53 $\rightarrow$ 53 & 0.65 $\rightarrow$ 0.65 & 90 $\rightarrow$ 57 & 0.05 $\rightarrow$ 0.04 \\ \bottomrule
    \end{tabular}
}
\caption{\textbf{Effectiveness of AT.} AT is ineffective against \attname{}. 
The defended models completely lose the computational efficiency (\textit{i.e.}, they have $\sim$0 efficacy), 
even with the aggressive setting with the patience of 2.
\textbf{P} is the patience.}
\label{tbl:at-results}
\vspace{-1.em}
\end{wraptable}

%% file: sections/conclusion.tex
%
\section{Conclusion}
\label{sec:conclusion}

This work shows that the computational savings 
that input-adaptive multi-exit language models offer 
are \emph{not} robust against adversarial slowdown.
To evaluate, we propose \attname{}, 
an adversarial text-crafting algorithm 
with the objective of bypassing early-exit points.
\attname{} significantly reduces the computational savings 
offered by those models.
More sophisticated input-adaptive mechanisms 
suited for higher efficacy 
become more vulnerable to slowdown attacks. 
Our linguistic analysis exposes 
that it is not about the magnitude of perturbations 
but because pushing an input outside the distribution 
on which a model is trained is easy.
We also show the limits of adversarial training in defeating our attacks 
and the effectiveness of input sanitization as a defense.
Our results suggest that future research is required
to develop efficient yet robust input-adaptive multi-exit inference.


%% file: sections/limitation.tex
\section{Limitations, Societal Impacts, and Future Work}
\label{sub:limitations-societal-impacts}

%
As shown in our work, word-level perturbations carefully calibrated by \attname{} make the resulting natural adversarial texts offset the computational savings multi-exit language models provide. However, there have been other types of text perturbations, e.g., character-level~\cite{CharAttack, BadChar} or sentence-level perturbations~\cite{T3}. We have not tested whether an adversary can adapt them to 
cause slowdowns. If these new attacks are successful, we can hypothesize that some other attributes of language models contribute to lowering the confidence of the predictions made by internal classifiers (early-exit points). It may also render potential countermeasures, such as input sanitization, ineffective. 
Future work is needed to investigate 
attacks exploiting different perturbations to cause adversarial slowdown.

%
To foster future research, we developed \attname{} in an open-source adversarial attack framework, TextAttack~\cite{TextAttack}. This will make our attacks more accessible to the community. A concern is that a potential adversary can use those attacks to push the behaviors of systems that harness multi-exit mechanisms outside the expectations. But we believe that our offensive measures will be adopted broadly by practitioners and have them audit such systems before they are publicly available.

%
We have also shown that using state-of-the-art conversational models, such as ChatGPT, to sanitize perturbed inputs can be an effective defense against adversarial slowdown. But it is unclear what attributes of those models were able to remove the artifacts (i.e., perturbations) our attack induces. Moreover, the fact that this defense heavily relies on the referencing model's capability that the victim cannot control may give room for an adversary to develop stronger attacks in the future. 


%
It is also possible that when using conversational models online as a potential countermeasure, there will be risks of data leakage. However, our proposal does not mean to use ChatGPT as-is. Instead, since other input sanitation (Grammarly) failed, we used it as an accessible tool for input sanitization via a conversational model as a proof-of-concept that it may have effectiveness as a defense. Alternatively, a defender can compose input sanitization as a completely in-house solution by leveraging off-the-shelf models like Vicuna-13B~\cite{chiang2023vicuna}. We leave this exploration as future work.

%
An interesting question is to what extent models like ChatGPT offer robustness to the \emph{conventional} adversarial attacks that aim to reduce a model's utility in the inference time. But this is not the scope of our work. While the conversational models we use offer some robustness to our slowdown attacks with fewer side-effects, it does not mean that this direction is bulletproof against all adversarial attacks and/or adaptive adversaries in the future. Recent work shows two opposite views about the robustness of conversational models~\cite{wang2023robustness, zou2023universal}. We envision more future work on this topic.

%
We find that the runtime of the potential countermeasures we explore in Sec~\ref{sec:defenses} is higher than the average inference time of \emph{undefended} multi-exit language models. This would make them useless from a pure runtime standpoint. However, we reason that the purpose of using these defenses was primarily exploratory, aiming to understand further why specific text causes more slowdown and how modifying such text can revert this slowdown. Moreover, input sanitization is already used in commercial models. Claude-2\footnote{\href{https://claude.ai}{https://claude.ai}}, a conversational model similar to ChatGPT, already employs input-filtering techniques, which we believe, when combined together, is a promising future work direction. Defenses with less computational overheads must be an important area for future work.

Overall, this work raises an open-question to the community about the feasibility of \emph{input-adaptive} efficient inference on large language models. We believe future work is necessary to evaluate this feasibility and develop a mechanism that kills two birds (efficacy and robustness) with one method.


%% file: sections/ack.tex
\section*{Acknowledgements}

We thank the anonymous reviewers for their constructive feedback.
Zachary Coalson and Sanghyun Hong are partially supported by 
the Google Faculty Research Award and the Samsung Global Research Outreach (GRO) program.
Gabriel Ritter and Rakesh Bobba are partially supported by the U.S. Department of Transportation. 
The findings and conclusions in this work are those of the author(s) and do not necessarily represent the views of the funding agency.
%


%% file: sections/appendix.tex
\appendix

\section{Comparison to Related Attacks in Prior Work}
\label{appendix:novelty}

Here, we expand upon our discussion in Sec~\ref{sec:related-work} and discuss the novelty of our slowdown attack compared to the attacks developed in prior work~\cite{BadChar, Sponge, DeepSloth}.

Prior work~\cite{BadChar, Sponge} has shown that an adversary can increase the energy consumption of language models in modern computing hardware via ``sponge'' examples. These inputs exploit computational properties of hardware or tokenization, e.g., input dimensionality and/or activation sparsity, to increase the inference runtime. In contrast, our attack is hardware-agnostic and targets multi-exit language models, a new algorithm for efficient language model computations. No prior work has been done on adversarial slowdowns in the context of multi-exit language models, and our attack is the first that generates natural adversarial text that bypasses the early-exit layers of multi-exit models.

Compared to the slowdown attacks in the computer vision domain~\cite{DeepSloth}, we also highlight the unique challenges we address: (1) Against language models, we often do not have access to input gradients (which is straightforward in attacks against computer vision models). We thus need to design a new slowdown objective compatible with non-gradient-based attacks. (2) We must bound the values of our slowdown objective within [0, 1]. The objective used in the prior work~\cite{DeepSloth} is unbounded to [0, $\infty$]; thus, a straightforward adaptation of this objective for adversarial text-attack algorithms leads to unbounded perturbations, and the resulting text completely differs from the original one. (3) The attack against language models works with discrete text inputs; not all embedding-level perturbations we compute exist as words, and small changes to input (word, characters) can result in large logit changes. We must search for candidate words (or word combinations) for substitution.

\section{Experimental Setup in Detail}
\label{appendix:setup-detail}

Here, we describe our experimental setup in detail.
We implement all the multi-exit mechanisms and our attacks
using Python v3.9\footnote{Python: \href{https://www.python.org}{https://www.python.org}}
and PyTorch v1.10\footnote{PyTorch: \href{https://pytorch.org}{https://pytorch.org}}
that supports CUDA 11.7 for accelerating computations by using GPUs.
We take the pre-trained language models
(\textit{i.e.}, BERT and ALBERT)
from Hugging Face\footnote{Hugging Face: \href{https://huggingface.co}{https://huggingface.co}}
and fine-tune them on GLUE benchmarks.
Our experiments run on a machine equipped with
Intel Xeon Processor with 48 cores,
64GB memory and 8 Nvidia A40 GPUs.

\topic{Multi-exit language models.}
All the early-exit mechanisms we employ, 
\textit{i.e.}, DeeBERT, PABEE, and PastFuture,
takes a pre-trained language model,
attaches an internal classifier (\textit{i.e.}, an early-exit) to each internal layer,
and fine-tune the entire model on a task.
We choose the pre-trained BERT (`bert-base-uncased')
and ALBERT (`albert-base-v2') from Hugging Face.
We fine-tune them on seven different GLUE tasks for five epochs.
We choose a batch-size from {32, 64, 128}
and a learning rate from {1e-5, 2e-5, 3e-5, 4e-5, 5e-5}.
We perform hyper-parameter sweeping over all the combinations
and select the models that provide the best accuracy for each task.
We select the early-exit thresholds 
based on the values in the original studies.
In DeeBERT~\cite{DeeBERT}, we pick the entropy
that offers 1.5$\times$ computational speedup.
In PABEE~\cite{PABEE}, we choose the patience value of 6.
In PastFuture~\cite{PastFuture}, we set the entropy values
where we achieve 2$\times$ speedup.

\topic{Choice of the slowdown metric.}
Prior work on early-exit mechanisms uses two metrics:
\emph{wall-clock time} and \emph{speedup}.
DeeBERT uses wall-clock time,
but it is not a desirable metric
as the metric depends on the choice of hardware or software libraries, 
such as deep learning frameworks.
PABEE and PastFuture propose speedup,
a ratio between the total number of layers
and the number of layers required to make a prediction.
They compute this ratio over the entire test-set samples
and report the average value.
However, it is also not an accurate estimation of the computational savings,
as depending on model architectures, 
the number of parameters in a layer
and the way it computes the inputs could be different. 
As a result, we employ \emph{efficacy}
that counts the number of floating-point computations.
%
Note that BERT and ALBERT are both stacks of Transformer layers; this, luckily, the speedup is the inverse of the efficacy.

\input{algorithms/a2t}

\section{The \attname{} Attack Based on A2T}
\label{appendix:waffle-a2t}

We show how we adapt A2T~\cite{A2T} 
for auditing the slowdown risk in Algorithm~\ref{alg:a2t}.
We highlighted our adaptation to A2T in {\color{blue}blue}.

\topic{(line 1--2) Compute word importance.}
We first compute the importance of each word $w_i$ in a text input $x$.
The procedure is the same as shown in Sec~\ref{subsec:our-attack};
we remove each word from $x$
and compute the influence on the slowdown objective.
We then rank the words based on how much each removal increases the slowdown score $s_i$.
We also filter out stop words, 
\textit{e.g.}, `the' or `when'.

\topic{(line 3--13) Craft a natural adversarial text.}
The attack then works by replacing a set of words
in $x$ with the candidates carefully chosen by $T(x^*, i)$.
The transformation function $T$
selects the top 20 synonyms
that has the similar embeddings~\cite{WordEmbedding},
based on the cosine similarity.
We only keep the candidates
with the same part-of-speech as $w_i$
to minimize grammar destruction.

We then substitute $w_i$ with the candidate 
that maximizes the slowdown score $s_i(x^t, f_{\theta})$
after the substitution.
If the text after this substitution $x^*$ 
increases the slowdown score over the threshold $\alpha$,
we return $x^*$.
However, when there is no such candidate,
we pick the candidate with the highest slowdown score,
substitute the candidate with $w_i$,
and repeat the same procedure with the next word $w_{i+1}$.
In the end, we return $x^*$
even when the slowdown score does not meet the threshold $\alpha$.

\section{Data and Code Availability}
\label{sec:data-and-models}

As a part of the reproducible research practice,
we release our data and source code along with our submission.
Our \attname{} attacks are implemented using TextAttack~\cite{TextAttack},
a Python framework for testing a model's robustness to adversarial attacks.
We also include our attacks on the TextAttack repo\footnote{TextAttack: \href{https://github.com/QData/TextAttack}{https://github.com/QData/TextAttack}}.
This will encourage practitioners and AI-system engineers
developing (or employing) input-adaptive efficient inference mechanisms
to test their robustness to adversarial slowdown.

%
\section{More Results on Transferability of \attname{}}
\label{sec:more-results-transferability}

Here we provide further results from our transferability experiments 
in Sec~\ref{sec:black-box-attacks}. For the cross-seed and cross-mechanism attacks, we show all victim-surrogate combinations involving all three early-exit mechanisms. For the cross-architecture attack, we show our results on all seven GLUE tasks.

\begin{table*}[ht]
\centering
\begin{threeparttable}
\adjustbox{max width=\textwidth}{
    \begin{tabular}{c|c|c||c|cc|cc}
    \toprule
    \multirow{2}{*}{\textbf{Model}} & \multirow{2}{*}{\textbf{Arch.}} & \multirow{2}{*}{\textbf{Mechanism}} & \multicolumn{1}{c|}{\multirow{2}{*}{\textbf{Type}}} & \multicolumn{2}{c|}{\textbf{RTE}} & \multicolumn{2}{c}{\textbf{QQP}} \\
     &  &  & \multicolumn{1}{c|}{} & \textbf{Acc.} & \textbf{Eff.} & \textbf{Acc.} & \textbf{Eff.} \\ \midrule \midrule
    \textbf{S} & \textbf{BERT} & \textbf{DeeBERT} & S $\rightarrow$ S & 67 $\rightarrow$ 55 & 0.32 $\rightarrow$ 0.11 & 91 $\rightarrow$ 76 & 0.32 $\rightarrow$ 0.18 \\
    \textbf{V} & \textbf{BERT} & \textbf{DeeBERT} & S $\rightarrow$ V & 64 $\rightarrow$ 52 & 0.36 $\rightarrow$ 0.23 & 91 $\rightarrow$ 77 & 0.35 $\rightarrow$ 0.30 \\ \midrule \midrule
    \textbf{S} & \textbf{BERT} & \textbf{PABEE} & S $\rightarrow$ S & 66 $\rightarrow$ 49 & 0.22 $\rightarrow$ 0.08 & 91 $\rightarrow$ 69 & 0.35 $\rightarrow$ 0.16 \\
    \textbf{V} & \textbf{BERT} & \textbf{PABEE} & S $\rightarrow$ V & 65 $\rightarrow$ 61 & 0.22 $\rightarrow$ 0.14 & 91 $\rightarrow$ 71 & 0.35 $\rightarrow$ 0.26 \\ \midrule \midrule
    \textbf{S} & \textbf{BERT} & \textbf{PastFuture} & S $\rightarrow$ S & 66 $\rightarrow$ 52 & 0.47 $\rightarrow$ 0.11 & 91 $\rightarrow$ 72 & 0.50 $\rightarrow$ 0.26 \\
    \textbf{V} & \textbf{BERT} & \textbf{PastFuture} & S $\rightarrow$ V & 66 $\rightarrow$ 55 & 0.50 $\rightarrow$ 0.25 & 91 $\rightarrow$ 72 & 0.51 $\rightarrow$ 0.33 \\ \bottomrule
    \end{tabular}
}
\begin{tablenotes}
    \item S = Surrogate model; V = Victim model
\end{tablenotes}
\end{threeparttable}
\caption{\textbf{Cross-seed attack results.} 
In all cases, the efficacy 
of the white-box attacks (S$\rightarrow$S) is significantly reduced while the efficacy of the transfer attacks (S$\rightarrow$V) comparatively drops.}
\label{tbl:cross-seed}
\end{table*}

Table~\ref{tbl:cross-seed} shows the entire results from the cross-seed transfer-based attacks. Examining all the three early-exit mechanisms, attacking the victim model using the adversarial texts crafted on the surrogate models causes approximately 50\% of the slowdown induced when attacking the surrogate directly. 

\begin{table*}[ht]
\centering
\begin{threeparttable}
\adjustbox{max width=\textwidth}{
    \begin{tabular}{c|c|c||c|cc|cc}
    \toprule
    \multirow{2}{*}{\textbf{Model}} & \multirow{2}{*}{\textbf{Arch.}} & \multirow{2}{*}{\textbf{Mechanism}} & \multirow{2}{*}{\textbf{Type}} & \multicolumn{2}{c|}{\textbf{RTE}} & \multicolumn{2}{c}{\textbf{QQP}} \\
     &  &  &  & \textbf{Acc.} & \textbf{Eff.} & \textbf{Acc.} & \textbf{Eff.} \\ \midrule \midrule
    \textbf{S} & \textbf{BERT} & \textbf{PastFuture} & S $\rightarrow$ S & 66 $\rightarrow$ 52 & 0.47 $\rightarrow$ 0.11 & 91 $\rightarrow$ 72 & 0.50 $\rightarrow$ 0.26 \\
    \textbf{V} & \textbf{BERT} & \textbf{DeeBERT} & S $\rightarrow$ V & 64 $\rightarrow$ 51 & 0.36 $\rightarrow$ 0.26 & 91 $\rightarrow$ 70 & 0.35 $\rightarrow$ 0.28 \\ \midrule
    \textbf{S} & \textbf{BERT} & \textbf{PABEE} & S $\rightarrow$ S & 66 $\rightarrow$ 49 & 0.22 $\rightarrow$ 0.08 & 91 $\rightarrow$ 69 & 0.35 $\rightarrow$ 0.16 \\
    \textbf{V} & \textbf{BERT} & \textbf{DeeBERT} & S $\rightarrow$ V & 64 $\rightarrow$ 50 & 0.36 $\rightarrow$ 0.28 & 91 $\rightarrow$ 73 & 0.35 $\rightarrow$ 0.31 \\ \midrule
    \textbf{S} & \textbf{BERT} & \textbf{DeeBERT} & S $\rightarrow$ S & 67 $\rightarrow$ 55 & 0.32 $\rightarrow$ 0.11 & 91 $\rightarrow$ 76 & 0.32 $\rightarrow$ 0.18 \\
    \textbf{V} & \textbf{BERT} & \textbf{PABEE} & S $\rightarrow$ V & 65 $\rightarrow$ 53 & 0.22 $\rightarrow$ 0.19 & 91 $\rightarrow$ 67 & 0.35 $\rightarrow$ 0.26 \\ \midrule
    \textbf{S} & \textbf{BERT} & \textbf{PastFuture} & S $\rightarrow$ S & 66 $\rightarrow$ 52 & 0.47 $\rightarrow$ 0.11 & 91 $\rightarrow$ 72 & 0.50 $\rightarrow$ 0.26 \\
    \textbf{V} & \textbf{BERT} & \textbf{PABEE} & S $\rightarrow$ V & 65 $\rightarrow$ 57 & 0.22 $\rightarrow$ 0.18 & 91 $\rightarrow$ 76 & 0.35 $\rightarrow$ 0.31 \\
    \midrule 
    \textbf{S} & \textbf{BERT} & \textbf{DeeBERT} & S $\rightarrow$ S & 67 $\rightarrow$ 55 & 0.32 $\rightarrow$ 0.11 & 91 $\rightarrow$ 76 & 0.32 $\rightarrow$ 0.18 \\
    \textbf{V} & \textbf{BERT} & \textbf{PastFuture} & S $\rightarrow$ V & 66 $\rightarrow$ 54 & 0.50 $\rightarrow$ 0.40 & 91 $\rightarrow$ 76 & 0.51 $\rightarrow$ 0.46 \\ \midrule 
    \textbf{S} & \textbf{BERT} & \textbf{PABEE} & S $\rightarrow$ S & 66 $\rightarrow$ 49 & 0.22 $\rightarrow$ 0.08 & 91 $\rightarrow$ 69 & 0.35 $\rightarrow$ 0.16 \\
    \textbf{V} & \textbf{BERT} & \textbf{PastFuture} & S $\rightarrow$ V & 66 $\rightarrow$ 55 & 0.50 $\rightarrow$ 0.29 & 91 $\rightarrow$ 74 & 0.51 $\rightarrow$ 0.38 \\ \bottomrule
    \end{tabular}
}
\begin{tablenotes}
    \item S = Surrogate model; V = Victim model
\end{tablenotes}
\end{threeparttable}
\caption{\textbf{Cross-mechanism attack results.} 
While not as effective as the cross-seed attack, marginal efficacy drops are seen for most victim-surrogate pairs.}
\label{tbl:cross-mechanism}
\end{table*}

Table~\ref{tbl:cross-mechanism} shows all results from the cross-mechanism attack scenario. In a majority of victim-surrogate combinations, we observe the slowdown similar to the cross-seed scenario.
This makes sense, as only the early-exit mechanisms differ which account for a small number of parameters relative to the entire model. 
The result also imply that even if an attacker does not know the specific early-exit mechanism of the target model, high slowdown can still be induced.

\begin{table*}[ht]
\centering
\begin{threeparttable}
\adjustbox{max width=\textwidth}{
   \begin{tabular}{c|c|c|c|c|c|c|c|c|c|c}
   \toprule
    \textbf{Model} & \textbf{Arch.} & \textbf{Type} & \textbf{Metric} & \textbf{RTE} & \textbf{MNLI} & \textbf{MRPC} & \textbf{QNLI} & \textbf{QQP} & \textbf{SST-2} & \textbf{CoLA} \\ \midrule \midrule
    \multirow{2}{*}{\textbf{S}} & \multirow{2}{*}{\textbf{ALBERT}} & \multirow{2}{*}{S $\rightarrow$ S} & \textbf{Acc.} & 77 $\rightarrow$ 55 & 85 $\rightarrow$ 30 & 88 $\rightarrow$ 49 & 90 $\rightarrow$ 55 & 91 $\rightarrow$ 71 & 92 $\rightarrow$ 56 & 81 $\rightarrow$ 49 \\
     &  &  & \textbf{Eff.} & 0.25 $\rightarrow$ 0.12 & 0.28 $\rightarrow$ 0.06 & 0.33 $\rightarrow$ 0.11 & 0.32 $\rightarrow$ 0.07 & 0.36 $\rightarrow$ 0.19 & 0.36 $\rightarrow$ 0.07 & 0.32 $\rightarrow$ 0.08 \\ \midrule
    \multirow{2}{*}{\textbf{V}} & \multirow{2}{*}{\textbf{BERT}} & \multirow{2}{*}{S $\rightarrow$ V} & \textbf{Acc.} & 65 $\rightarrow$ 51 & 83 $\rightarrow$ 24 & 82 $\rightarrow$ 43 & 89 $\rightarrow$ 69 & 91 $\rightarrow$ 78 & 91 $\rightarrow$ 71 & 82 $\rightarrow$ 59 \\
     &  &  & \textbf{Eff.} & 0.22 $\rightarrow$ 0.20 & 0.24 $\rightarrow$ 0.16 & 0.34 $\rightarrow$ 0.10 & 0.29 $\rightarrow$ 0.24 & 0.35 $\rightarrow$ 0.34 & 0.32 $\rightarrow$ 0.21 & 0.34 $\rightarrow$ 0.24 \\ \midrule \midrule
    \multirow{2}{*}{\textbf{S}} & \multirow{2}{*}{\textbf{BERT}} & \multirow{2}{*}{S $\rightarrow$ S} & \textbf{Acc.} & 66 $\rightarrow$ 49 & 82 $\rightarrow$ 30 & 81 $\rightarrow$ 54 & 89 $\rightarrow$ 57 & 91 $\rightarrow$ 69 & 91 $\rightarrow$ 55 & 77 $\rightarrow$ 63 \\
     &  &  & \textbf{Eff.} & 22 $\rightarrow$ 0.08 & 0.24 $\rightarrow$ 0.06 & 0.33 $\rightarrow$ 0.14 & 0.29 $\rightarrow$ 0.08 & 0.35 $\rightarrow$ 0.16 & 0.32 $\rightarrow$ 0.07 & 0.37 $\rightarrow$ 0.26 \\ \midrule
    \multirow{2}{*}{\textbf{V}} & \multirow{2}{*}{\textbf{ALBERT}} & \multirow{2}{*}{S $\rightarrow$ V} & \textbf{Acc.} & 77 $\rightarrow$ 63 & 85 $\rightarrow$ 22 & 86 $\rightarrow$ 81 & 91 $\rightarrow$ 74 & 91 $\rightarrow$ 74 & 92 $\rightarrow$ 68 & 81 $\rightarrow$ 47 \\
     &  &  & \textbf{Eff.} & 0.22 $\rightarrow$ 0.21 & 0.28 $\rightarrow$ 0.16 & 0.32 $\rightarrow$ 0.22 & 0.32 $\rightarrow$ 0.22 & 0.36 $\rightarrow$ 0.34 & 0.36 $\rightarrow$ 0.28 & 0.22 $\rightarrow$ 0.22 \\ \bottomrule
    \end{tabular}
}
\begin{tablenotes}
    \item S = Surrogate model; V = Victim model
\end{tablenotes}
\end{threeparttable}
\caption{\textbf{Cross-architecture attack results.} 
With PABEE as the early-exit mechanism, we attack BERT-based models with an ALBERT-based surrogate and vice-versa on all GLUE tasks.}
\label{tbl:cross-architecture}
\end{table*}

Table~\ref{tbl:cross-architecture} shows all results from the cross-architecture attack. 
We run our experiments with the entire GLUE tasks. Compared to the previous two attacks, the cross-architecture attack 
is less effective. This implies that knowing the target's architecture is a critical 
when 
exploiting adversarial transferability. If architecture is known, a strong attack can still be launched even if the early-exit mechanism and parameter values remain unknown to the attacker.

%
\section{More Discussion on Our Linguistic Analysis Results}
\label{sec:more-linguistic-analysis}

Here, we provide further insights regarding our linguistic analysis performed in Sec~\ref{sec:qualitative-study}. Conventional wisdom from studies in computer vision suggests that if an adversary leverages larger input perturbations (e.g., the perturbations are bounded to 16 pixels), their attack will be stronger than attacks with smaller input perturbations (e.g., 8 pixels). In other words, if a model is robust against attacks perturbing 16 pixels at most, the model is also robust to the 8-pixel bounded perturbations.

However, we find that this is \emph{not true} for our slowdown attacks. Investigating the adversarial texts generated from our ``unbounded'' slowdown attacks, we could not find the correlation between the attack strength and the number of word-level perturbations. This result questions the effectiveness of adversarial training (AT), a standard defense that trains a model with bounded adversarial texts~\cite{miyato2017adversarial, zhu2019freelb, jiang2019smart, liu2020adversarial, A2T}. 
In Sec~\ref{sec:defenses}, we show that vanilla AT is an ineffective countermeasure (and also causes undesirable consequences, e.g., the utility and efficacy loss of a model). 

We also offer an alternative insight for developing future defenses. In Sec~\ref{sec:qualitative-study}, we show that an adversary can exploit the subject-predicate mismatch to make a model less confident about the perturbed sample's prediction. This misalignment, while easier for humans to identify, is difficult for a target model to do so. Thus, in Sec~\ref{sec:defenses}, we propose to leverage models able to correct grammatical errors, including the subject-predicate mismatches, for sanitizing inputs before being fed to the target multi-exit models. But we find that such models are either far too slow to be practical or do not offer enough benefits. The result suggests future work in input sanitization for fast and effective methods. 

%
\section{Impact of \attname{} on Runtime}
\label{sec:runtime-analysis}

We provide results on the impact of our attacks on \emph{runtime} and compare it with the efficacy metric we use. 
Table~\ref{tbl:runtime-results} 
shows our results on DeeBERT across multiple datasets.

\begin{table}[ht]
\adjustbox{max width=\textwidth}{
    \begin{tabular}{c|cc|cc|cc|cc|cc}
    \toprule
    \multirow{2}{*}{\textbf{Metric}} & \multicolumn{2}{c|}{\textbf{QQP}} & \multicolumn{2}{c|}{\textbf{RTE}} & \multicolumn{2}{c|}{\textbf{QNLI}} & \multicolumn{2}{c|}{\textbf{MRPC}} & \multicolumn{2}{c}{\textbf{CoLA}} \\
                                     & \textbf{Clean}  & \textbf{WAFFLE} & \textbf{Clean}  & \textbf{WAFFLE} & \textbf{Clean}  & \textbf{WAFFLE}  & \textbf{Clean}  & \textbf{WAFFLE}  & \textbf{Clean}  & \textbf{WAFFLE}  \\ \midrule \midrule
    \textbf{Efficacy}                & 0.36            & 0.22            & 0.34            & 0.12            & 0.35            & 0.09             & 0.35            & 0.10             & 0.34            & 0.13             \\
    \textbf{Runtime}                 & 7.50s           & 9.09s            & 2.75s           & 3.41s            & 3.73s            & 4.68s             & 7.78s            & 10.70s            & 7.84s            & 10.04s            \\ \hline
    \end{tabular}
}
\caption{\textbf{Impact of \attname{} on runtime.}
With DeeBERT as the victim's mechanism, we report the runtime (in seconds) and efficacy of clean and perturbed samples on five GLUE tasks. We run our experiments on a single Tesla V100 GPU. These results indicate that \attname{} increases the actual runtime of multi-exit language models and that runtime is inversely correlated to efficacy.}
\label{tbl:runtime-results}
\end{table}

The results show that \attname{} increases the actual runtime of multi-exit language models, i.e., our slowdown results apply to real-world scenarios. Additionally, a reduction in efficacy is correlated with an increase in runtime. We choose efficacy as a metric to quantify the slowdown (as opposed to runtime) because the metric is hardware agnostic. The exit layer number we use to compute efficacy will not change between models run on different hardware configurations. 

%% file: algorithms/a2t.tex
\begin{wrapfigure}{R}{0.54\linewidth}
\begin{minipage}{\linewidth}

\vspace{-1.1em}     

\begin{algorithm}[H]
\caption{\attname{} (based on A2T)}
\label{alg:a2t}
\textbf{Input:} a text input $x = \{w_1, w_2, ..., w_n\}$; the victim model $f_{\theta}$; a transformation module $T(x, i)$ that perturbs $x$ by replacing $w_i$; and {\color{blue}the success threshold $\alpha$}. \\
\textbf{Output:} a natural adversarial text $x^*$
\begin{algorithmic}[1]
    \State {\color{blue}Calculate $I(w_i)$} for all words $w_i$ by making one forward and backward pass.
    
    \State $R \gets$ ranking $r_1, r_2, ..., r_n$ of words $w_1, ..., w_n$ by descending importance
    \State $x^* \gets x$
    \For{$i = r_1, r_2, ..., r_n$ in $R$}
        \State $X_{cand} \gets T(x^*, i)$
        \If{$X_{cand} \neq \emptyset$}
            \State $x^* \gets \argmax_{x^{t}\in X_{cand}}$ {\color{blue}$s_i(x^{t}, f_{\theta})$}
            \If{{\color{blue}$s_i \geq \alpha$}}
                \State \Return{$x^*$}
            \EndIf
        \EndIf
    \EndFor
    \State {\color{blue}\Return $x^*$}
\end{algorithmic}
\end{algorithm}

\vspace{-3.4em} 

\end{minipage}
\end{wrapfigure}

%% file: neurips_2023.bbl
\begin{thebibliography}{49}
\providecommand{\natexlab}[1]{#1}
\providecommand{\url}[1]{\texttt{#1}}
\expandafter\ifx\csname urlstyle\endcsname\relax
  \providecommand{\doi}[1]{doi: #1}\else
  \providecommand{\doi}{doi: \begingroup \urlstyle{rm}\Url}\fi

\bibitem[Boucher et~al.(2022)Boucher, Shumailov, Anderson, and
  Papernot]{BadChar}
N.~Boucher, I.~Shumailov, R.~Anderson, and N.~Papernot.
\newblock Bad characters: Imperceptible nlp attacks.
\newblock In \emph{2022 IEEE Symposium on Security and Privacy (SP)}, pages
  1987--2004. IEEE, 2022.

\bibitem[Carlini and Wagner(2017)]{CWAttack}
N.~Carlini and D.~Wagner.
\newblock Towards evaluating the robustness of neural networks.
\newblock In \emph{2017 IEEE Symposium on Security and Privacy (SP)}, pages
  39--57, 2017.
\newblock \doi{10.1109/SP.2017.49}.

\bibitem[Cer et~al.(2018)Cer, Yang, Kong, Hua, Limtiaco, St.~John, Constant,
  Guajardo-Cespedes, Yuan, Tar, Strope, and Kurzweil]{USE}
D.~Cer, Y.~Yang, S.-y. Kong, N.~Hua, N.~Limtiaco, R.~St.~John, N.~Constant,
  M.~Guajardo-Cespedes, S.~Yuan, C.~Tar, B.~Strope, and R.~Kurzweil.
\newblock Universal sentence encoder for {E}nglish.
\newblock In \emph{Proceedings of the 2018 Conference on Empirical Methods in
  Natural Language Processing: System Demonstrations}, pages 169--174,
  Brussels, Belgium, Nov. 2018. Association for Computational Linguistics.
\newblock \doi{10.18653/v1/D18-2029}.
\newblock URL \url{https://aclanthology.org/D18-2029}.

\bibitem[Chiang et~al.(2023)Chiang, Li, Lin, Sheng, Wu, Zhang, Zheng, Zhuang,
  Zhuang, Gonzalez, et~al.]{chiang2023vicuna}
W.-L. Chiang, Z.~Li, Z.~Lin, Y.~Sheng, Z.~Wu, H.~Zhang, L.~Zheng, S.~Zhuang,
  Y.~Zhuang, J.~E. Gonzalez, et~al.
\newblock Vicuna: An open-source chatbot impressing gpt-4 with 90\%* chatgpt
  quality.
\newblock \emph{See https://vicuna. lmsys. org (accessed 14 April 2023)}, 2023.

\bibitem[Devlin et~al.(2019)Devlin, Chang, Lee, and Toutanova]{BERT}
J.~Devlin, M.-W. Chang, K.~Lee, and K.~Toutanova.
\newblock Bert: Pre-training of deep bidirectional transformers for language
  understanding, 2019.

\bibitem[Ebrahimi et~al.(2018{\natexlab{a}})Ebrahimi, Lowd, and
  Dou]{CharAttack}
J.~Ebrahimi, D.~Lowd, and D.~Dou.
\newblock On adversarial examples for character-level neural machine
  translation.
\newblock \emph{CoRR}, abs/1806.09030, 2018{\natexlab{a}}.
\newblock URL \url{http://arxiv.org/abs/1806.09030}.

\bibitem[Ebrahimi et~al.(2018{\natexlab{b}})Ebrahimi, Rao, Lowd, and
  Dou]{HotFlip}
J.~Ebrahimi, A.~Rao, D.~Lowd, and D.~Dou.
\newblock {H}ot{F}lip: White-box adversarial examples for text classification.
\newblock In \emph{Proceedings of the 56th Annual Meeting of the Association
  for Computational Linguistics (Volume 2: Short Papers)}, pages 31--36,
  Melbourne, Australia, July 2018{\natexlab{b}}. Association for Computational
  Linguistics.
\newblock \doi{10.18653/v1/P18-2006}.
\newblock URL \url{https://aclanthology.org/P18-2006}.

\bibitem[Garg and Ramakrishnan(2020)]{BAE20}
S.~Garg and G.~Ramakrishnan.
\newblock Bae: Bert-based adversarial examples for text classification, 2020.

\bibitem[Guo et~al.(2021)Guo, Sablayrolles, J{\'e}gou, and Kiela]{GDBA:21}
C.~Guo, A.~Sablayrolles, H.~J{\'e}gou, and D.~Kiela.
\newblock Gradient-based adversarial attacks against text transformers.
\newblock In \emph{Proceedings of the 2021 Conference on Empirical Methods in
  Natural Language Processing}, pages 5747--5757, Online and Punta Cana,
  Dominican Republic, Nov. 2021. Association for Computational Linguistics.
\newblock \doi{10.18653/v1/2021.emnlp-main.464}.
\newblock URL \url{https://aclanthology.org/2021.emnlp-main.464}.

\bibitem[Hong et~al.(2021)Hong, Kaya, Modoranu, and Dumitras]{DeepSloth}
S.~Hong, Y.~Kaya, I.-V. Modoranu, and T.~Dumitras.
\newblock A panda? no, it's a sloth: Slowdown attacks on adaptive multi-exit
  neural network inference.
\newblock In \emph{International Conference on Learning Representations}, 2021.
\newblock URL \url{https://openreview.net/forum?id=9xC2tWEwBD}.

\bibitem[Huang et~al.(2017)Huang, Chen, Li, Wu, Van Der~Maaten, and
  Weinberger]{MSDNets}
G.~Huang, D.~Chen, T.~Li, F.~Wu, L.~Van Der~Maaten, and K.~Q. Weinberger.
\newblock Multi-scale dense networks for resource efficient image
  classification.
\newblock \emph{arXiv preprint arXiv:1703.09844}, 2017.

\bibitem[Ivgi and Berant(2021)]{AT:NLP}
M.~Ivgi and J.~Berant.
\newblock Achieving model robustness through discrete adversarial training.
\newblock \emph{arXiv preprint arXiv:2104.05062}, 2021.

\bibitem[Jia et~al.(2019)Jia, Raghunathan, G{\"o}ksel, and
  Liang]{CertifiedRobustness}
R.~Jia, A.~Raghunathan, K.~G{\"o}ksel, and P.~Liang.
\newblock Certified robustness to adversarial word substitutions.
\newblock In \emph{Proceedings of the 2019 Conference on Empirical Methods in
  Natural Language Processing and the 9th International Joint Conference on
  Natural Language Processing (EMNLP-IJCNLP)}, pages 4129--4142, Hong Kong,
  China, Nov. 2019. Association for Computational Linguistics.
\newblock \doi{10.18653/v1/D19-1423}.
\newblock URL \url{https://aclanthology.org/D19-1423}.

\bibitem[Jiang et~al.(2019)Jiang, He, Chen, Liu, Gao, and Zhao]{jiang2019smart}
H.~Jiang, P.~He, W.~Chen, X.~Liu, J.~Gao, and T.~Zhao.
\newblock Smart: Robust and efficient fine-tuning for pre-trained natural
  language models through principled regularized optimization.
\newblock \emph{arXiv preprint arXiv:1911.03437}, 2019.

\bibitem[Jiang et~al.(2020)Jiang, He, Chen, Liu, Gao, and Zhao]{SMART}
H.~Jiang, P.~He, W.~Chen, X.~Liu, J.~Gao, and T.~Zhao.
\newblock {SMART}: Robust and efficient fine-tuning for pre-trained natural
  language models through principled regularized optimization.
\newblock In \emph{Proceedings of the 58th Annual Meeting of the Association
  for Computational Linguistics}, pages 2177--2190, Online, July 2020.
  Association for Computational Linguistics.
\newblock \doi{10.18653/v1/2020.acl-main.197}.
\newblock URL \url{https://aclanthology.org/2020.acl-main.197}.

\bibitem[Jin et~al.(2020)Jin, Jin, Zhou, and Szolovits]{TextFooler}
D.~Jin, Z.~Jin, J.~T. Zhou, and P.~Szolovits.
\newblock Is bert really robust? a strong baseline for natural language attack
  on text classification and entailment, 2020.

\bibitem[Kaya et~al.(2019)Kaya, Hong, and Dumitras]{SDNOverthink}
Y.~Kaya, S.~Hong, and T.~Dumitras.
\newblock Shallow-deep networks: Understanding and mitigating network
  overthinking.
\newblock In \emph{International conference on machine learning}, pages
  3301--3310. PMLR, 2019.

\bibitem[Lan et~al.(2020)Lan, Chen, Goodman, Gimpel, Sharma, and
  Soricut]{ALBERT}
Z.~Lan, M.~Chen, S.~Goodman, K.~Gimpel, P.~Sharma, and R.~Soricut.
\newblock Albert: A lite bert for self-supervised learning of language
  representations.
\newblock In \emph{International Conference on Learning Representations}, 2020.
\newblock URL \url{https://openreview.net/forum?id=H1eA7AEtvS}.

\bibitem[Li et~al.(2021)Li, Zhang, Peng, Chen, Brockett, Sun, and
  Dolan]{CLARE:21}
D.~Li, Y.~Zhang, H.~Peng, L.~Chen, C.~Brockett, M.-T. Sun, and B.~Dolan.
\newblock Contextualized perturbation for textual adversarial attack.
\newblock In \emph{Proceedings of the 2021 Conference of the North American
  Chapter of the Association for Computational Linguistics: Human Language
  Technologies}, pages 5053--5069, Online, June 2021. Association for
  Computational Linguistics.
\newblock \doi{10.18653/v1/2021.naacl-main.400}.
\newblock URL \url{https://aclanthology.org/2021.naacl-main.400}.

\bibitem[Li et~al.(2019)Li, Ji, Du, Li, and Wang]{TextBugger}
J.~Li, S.~Ji, T.~Du, B.~Li, and T.~Wang.
\newblock Textbugger: Generating adversarial text against real-world
  applications.
\newblock In \emph{26th Annual Network and Distributed System Security
  Symposium, {NDSS} 2019, San Diego, California, USA, February 24-27, 2019}.
  The Internet Society, 2019.
\newblock URL
  \url{https://www.ndss-symposium.org/ndss-paper/textbugger-generating-adversarial-text-against-real-world-applications/}.

\bibitem[Liao et~al.(2021)Liao, Zhang, Ren, Su, Sun, and He]{PastFuture}
K.~Liao, Y.~Zhang, X.~Ren, Q.~Su, X.~Sun, and B.~He.
\newblock A global past-future early exit method for accelerating inference of
  pre-trained language models.
\newblock In \emph{Proceedings of the 2021 Conference of the North American
  Chapter of the Association for Computational Linguistics: Human Language
  Technologies}, pages 2013--2023, Online, June 2021. Association for
  Computational Linguistics.
\newblock \doi{10.18653/v1/2021.naacl-main.162}.
\newblock URL \url{https://aclanthology.org/2021.naacl-main.162}.

\bibitem[Liu et~al.(2020{\natexlab{a}})Liu, Zhou, Wang, Zhao, Deng, and
  Ju]{FastBERT}
W.~Liu, P.~Zhou, Z.~Wang, Z.~Zhao, H.~Deng, and Q.~Ju.
\newblock {F}ast{BERT}: a self-distilling {BERT} with adaptive inference time.
\newblock In \emph{Proceedings of the 58th Annual Meeting of the Association
  for Computational Linguistics}, pages 6035--6044, Online, July
  2020{\natexlab{a}}. Association for Computational Linguistics.
\newblock \doi{10.18653/v1/2020.acl-main.537}.
\newblock URL \url{https://aclanthology.org/2020.acl-main.537}.

\bibitem[Liu et~al.(2020{\natexlab{b}})Liu, Cheng, He, Chen, Wang, Poon, and
  Gao]{ALUM}
X.~Liu, H.~Cheng, P.~He, W.~Chen, Y.~Wang, H.~Poon, and J.~Gao.
\newblock Adversarial training for large neural language models,
  2020{\natexlab{b}}.
\newblock URL \url{https://arxiv.org/abs/2004.08994}.

\bibitem[Liu et~al.(2020{\natexlab{c}})Liu, Cheng, He, Chen, Wang, Poon, and
  Gao]{liu2020adversarial}
X.~Liu, H.~Cheng, P.~He, W.~Chen, Y.~Wang, H.~Poon, and J.~Gao.
\newblock Adversarial training for large neural language models.
\newblock \emph{arXiv preprint arXiv:2004.08994}, 2020{\natexlab{c}}.

\bibitem[Madry et~al.(2018)Madry, Makelov, Schmidt, Tsipras, and Vladu]{PGD}
A.~Madry, A.~Makelov, L.~Schmidt, D.~Tsipras, and A.~Vladu.
\newblock Towards deep learning models resistant to adversarial attacks.
\newblock In \emph{International Conference on Learning Representations}, 2018.
\newblock URL \url{https://openreview.net/forum?id=rJzIBfZAb}.

\bibitem[Miyato et~al.(2017)Miyato, Dai, and Goodfellow]{miyato2017adversarial}
T.~Miyato, A.~M. Dai, and I.~Goodfellow.
\newblock Adversarial training methods for semi-supervised text classification.
\newblock In \emph{International Conference on Learning Representations}, 2017.
\newblock URL \url{https://openreview.net/forum?id=r1X3g2_xl}.

\bibitem[Morris et~al.(2020)Morris, Lifland, Yoo, Grigsby, Jin, and
  Qi]{TextAttack}
J.~Morris, E.~Lifland, J.~Y. Yoo, J.~Grigsby, D.~Jin, and Y.~Qi.
\newblock Textattack: A framework for adversarial attacks, data augmentation,
  and adversarial training in nlp.
\newblock In \emph{Proceedings of the 2020 Conference on Empirical Methods in
  Natural Language Processing: System Demonstrations}, pages 119--126, 2020.

\bibitem[Mrk{\v{s}}i{\'c} et~al.(2016)Mrk{\v{s}}i{\'c}, {\'O}~S{\'e}aghdha,
  Thomson, Ga{\v{s}}i{\'c}, Rojas-Barahona, Su, Vandyke, Wen, and
  Young]{WordEmbedding}
N.~Mrk{\v{s}}i{\'c}, D.~{\'O}~S{\'e}aghdha, B.~Thomson, M.~Ga{\v{s}}i{\'c},
  L.~M. Rojas-Barahona, P.-H. Su, D.~Vandyke, T.-H. Wen, and S.~Young.
\newblock Counter-fitting word vectors to linguistic constraints.
\newblock In \emph{Proceedings of the 2016 Conference of the North {A}merican
  Chapter of the Association for Computational Linguistics: Human Language
  Technologies}, pages 142--148, San Diego, California, June 2016. Association
  for Computational Linguistics.
\newblock \doi{10.18653/v1/N16-1018}.
\newblock URL \url{https://aclanthology.org/N16-1018}.

\bibitem[Raffel et~al.(2020)Raffel, Shazeer, Roberts, Lee, Narang, Matena,
  Zhou, Li, and Liu]{T5}
C.~Raffel, N.~Shazeer, A.~Roberts, K.~Lee, S.~Narang, M.~Matena, Y.~Zhou,
  W.~Li, and P.~J. Liu.
\newblock Exploring the limits of transfer learning with a unified text-to-text
  transformer.
\newblock \emph{Journal of Machine Learning Research}, 21\penalty0
  (140):\penalty0 1--67, 2020.
\newblock URL \url{http://jmlr.org/papers/v21/20-074.html}.

\bibitem[Ren et~al.(2019)Ren, Deng, He, and Che]{GeneticAdv}
S.~Ren, Y.~Deng, K.~He, and W.~Che.
\newblock Generating natural language adversarial examples through probability
  weighted word saliency.
\newblock In \emph{Proceedings of the 57th Annual Meeting of the Association
  for Computational Linguistics}, pages 1085--1097, Florence, Italy, July 2019.
  Association for Computational Linguistics.
\newblock \doi{10.18653/v1/P19-1103}.
\newblock URL \url{https://aclanthology.org/P19-1103}.

\bibitem[Rogers et~al.(2020)Rogers, Kovaleva, and
  Rumshisky]{Interpret:Bertology}
A.~Rogers, O.~Kovaleva, and A.~Rumshisky.
\newblock A primer in bertology: What we know about how bert works.
\newblock \emph{Transactions of the Association for Computational Linguistics},
  8:\penalty0 842--866, 2020.

\bibitem[Schwartz et~al.(2020)Schwartz, Stanovsky, Swayamdipta, Dodge, and
  Smith]{RTRJ:20}
R.~Schwartz, G.~Stanovsky, S.~Swayamdipta, J.~Dodge, and N.~A. Smith.
\newblock The right tool for the job: Matching model and instance complexities.
\newblock In \emph{Proceedings of the 58th Annual Meeting of the Association
  for Computational Linguistics}, pages 6640--6651, Online, July 2020.
  Association for Computational Linguistics.
\newblock \doi{10.18653/v1/2020.acl-main.593}.
\newblock URL \url{https://aclanthology.org/2020.acl-main.593}.

\bibitem[Shumailov et~al.(2021)Shumailov, Zhao, Bates, Papernot, Mullins, and
  Anderson]{Sponge}
I.~Shumailov, Y.~Zhao, D.~Bates, N.~Papernot, R.~Mullins, and R.~Anderson.
\newblock Sponge examples: Energy-latency attacks on neural networks, 2021.

\bibitem[Szegedy et~al.(2014)Szegedy, Zaremba, Sutskever, Bruna, Erhan,
  Goodfellow, and Fergus]{InitialAML}
C.~Szegedy, W.~Zaremba, I.~Sutskever, J.~Bruna, D.~Erhan, I.~J. Goodfellow, and
  R.~Fergus.
\newblock Intriguing properties of neural networks.
\newblock In Y.~Bengio and Y.~LeCun, editors, \emph{2nd International
  Conference on Learning Representations, {ICLR} 2014, Banff, AB, Canada, April
  14-16, 2014, Conference Track Proceedings}, 2014.
\newblock URL \url{http://arxiv.org/abs/1312.6199}.

\bibitem[Teerapittayanon et~al.(2016)Teerapittayanon, McDanel, and
  Kung]{BranchyNet}
S.~Teerapittayanon, B.~McDanel, and H.-T. Kung.
\newblock Branchynet: Fast inference via early exiting from deep neural
  networks.
\newblock In \emph{2016 23rd International Conference on Pattern Recognition
  (ICPR)}, pages 2464--2469. IEEE, 2016.

\bibitem[Wang et~al.(2019)Wang, Singh, Michael, Hill, Levy, and Bowman]{GLUE}
A.~Wang, A.~Singh, J.~Michael, F.~Hill, O.~Levy, and S.~R. Bowman.
\newblock {GLUE}: A multi-task benchmark and analysis platform for natural
  language understanding.
\newblock In \emph{International Conference on Learning Representations}, 2019.
\newblock URL \url{https://openreview.net/forum?id=rJ4km2R5t7}.

\bibitem[Wang et~al.(2020)Wang, Pei, Pan, Chen, Wang, and Li]{T3}
B.~Wang, H.~Pei, B.~Pan, Q.~Chen, S.~Wang, and B.~Li.
\newblock T3: Tree-autoencoder constrained adversarial text generation for
  targeted attack.
\newblock In \emph{Proceedings of the 2020 Conference on Empirical Methods in
  Natural Language Processing (EMNLP)}, pages 6134--6150, Online, Nov. 2020.
  Association for Computational Linguistics.
\newblock \doi{10.18653/v1/2020.emnlp-main.495}.
\newblock URL \url{https://aclanthology.org/2020.emnlp-main.495}.

\bibitem[Wang et~al.(2021)Wang, Xu, Wang, Gan, Cheng, Gao, Awadallah, and
  Li]{AdvGLUE}
B.~Wang, C.~Xu, S.~Wang, Z.~Gan, Y.~Cheng, J.~Gao, A.~H. Awadallah, and B.~Li.
\newblock Adversarial {GLUE}: A multi-task benchmark for robustness evaluation
  of language models.
\newblock In \emph{Thirty-fifth Conference on Neural Information Processing
  Systems Datasets and Benchmarks Track (Round 2)}, 2021.
\newblock URL \url{https://openreview.net/forum?id=GF9cSKI3A_q}.

\bibitem[Wang et~al.(2023)Wang, Hu, Hou, Chen, Zheng, Wang, Yang, Huang, Ye,
  Geng, et~al.]{wang2023robustness}
J.~Wang, X.~Hu, W.~Hou, H.~Chen, R.~Zheng, Y.~Wang, L.~Yang, H.~Huang, W.~Ye,
  X.~Geng, et~al.
\newblock On the robustness of chatgpt: An adversarial and out-of-distribution
  perspective.
\newblock \emph{arXiv preprint arXiv:2302.12095}, 2023.

\bibitem[Xie et~al.(2021)Xie, Lu, Wang, and Wang]{ELBERT:21}
K.~Xie, S.~Lu, M.~Wang, and Z.~Wang.
\newblock Elbert: Fast albert with confidence-window based early exit.
\newblock In \emph{ICASSP 2021 - 2021 IEEE International Conference on
  Acoustics, Speech and Signal Processing (ICASSP)}, pages 7713--7717, 2021.
\newblock \doi{10.1109/ICASSP39728.2021.9414572}.

\bibitem[Xin et~al.(2020)Xin, Tang, Lee, Yu, and Lin]{DeeBERT}
J.~Xin, R.~Tang, J.~Lee, Y.~Yu, and J.~Lin.
\newblock {D}ee{BERT}: Dynamic early exiting for accelerating {BERT} inference.
\newblock In \emph{Proceedings of the 58th Annual Meeting of the Association
  for Computational Linguistics}, pages 2246--2251, Online, July 2020.
  Association for Computational Linguistics.
\newblock URL \url{https://www.aclweb.org/anthology/2020.acl-main.204}.

\bibitem[Xin et~al.(2021)Xin, Tang, Yu, and Lin]{BERxIT:21}
J.~Xin, R.~Tang, Y.~Yu, and J.~Lin.
\newblock {BER}xi{T}: Early exiting for {BERT} with better fine-tuning and
  extension to regression.
\newblock In \emph{Proceedings of the 16th Conference of the European Chapter
  of the Association for Computational Linguistics: Main Volume}, pages
  91--104, Online, Apr. 2021. Association for Computational Linguistics.
\newblock \doi{10.18653/v1/2021.eacl-main.8}.
\newblock URL \url{https://aclanthology.org/2021.eacl-main.8}.

\bibitem[Yoo and Qi(2021)]{A2T}
J.~Y. Yoo and Y.~Qi.
\newblock Towards improving adversarial training of {NLP} models.
\newblock In \emph{Findings of the Association for Computational Linguistics:
  EMNLP 2021}, pages 945--956, Punta Cana, Dominican Republic, Nov. 2021.
  Association for Computational Linguistics.
\newblock \doi{10.18653/v1/2021.findings-emnlp.81}.
\newblock URL \url{https://aclanthology.org/2021.findings-emnlp.81}.

\bibitem[Zhang et~al.(2022)Zhang, Zhu, Zhang, Wang, Jin, and Chung]{PCEE:22}
Z.~Zhang, W.~Zhu, J.~Zhang, P.~Wang, R.~Jin, and T.-S. Chung.
\newblock {PCEE}-{BERT}: Accelerating {BERT} inference via patient and
  confident early exiting.
\newblock In \emph{Findings of the Association for Computational Linguistics:
  NAACL 2022}, pages 327--338, Seattle, United States, July 2022. Association
  for Computational Linguistics.
\newblock \doi{10.18653/v1/2022.findings-naacl.25}.
\newblock URL \url{https://aclanthology.org/2022.findings-naacl.25}.

\bibitem[Zhou et~al.(2020)Zhou, Xu, Ge, McAuley, Xu, and Wei]{PABEE}
W.~Zhou, C.~Xu, T.~Ge, J.~McAuley, K.~Xu, and F.~Wei.
\newblock Bert loses patience: Fast and robust inference with early exit.
\newblock In \emph{Advances in Neural Information Processing Systems},
  volume~33, pages 18330--18341. Curran Associates, Inc., 2020.
\newblock URL
  \url{https://proceedings.neurips.cc/paper/2020/file/d4dd111a4fd973394238aca5c05bebe3-Paper.pdf}.

\bibitem[Zhu et~al.(2019)Zhu, Cheng, Gan, Sun, Goldstein, and
  Liu]{zhu2019freelb}
C.~Zhu, Y.~Cheng, Z.~Gan, S.~Sun, T.~Goldstein, and J.~Liu.
\newblock Freelb: Enhanced adversarial training for natural language
  understanding.
\newblock \emph{arXiv preprint arXiv:1909.11764}, 2019.

\bibitem[Zhu et~al.(2020)Zhu, Cheng, Gan, Sun, Goldstein, and Liu]{FreeLB}
C.~Zhu, Y.~Cheng, Z.~Gan, S.~Sun, T.~Goldstein, and J.~Liu.
\newblock Freelb: Enhanced adversarial training for natural language
  understanding.
\newblock In \emph{International Conference on Learning Representations}, 2020.
\newblock URL \url{https://openreview.net/forum?id=BygzbyHFvB}.

\bibitem[Zhu(2021)]{LeeBERT:21}
W.~Zhu.
\newblock {L}ee{BERT}: Learned early exit for {BERT} with cross-level
  optimization.
\newblock In \emph{Proceedings of the 59th Annual Meeting of the Association
  for Computational Linguistics and the 11th International Joint Conference on
  Natural Language Processing (Volume 1: Long Papers)}, pages 2968--2980,
  Online, Aug. 2021. Association for Computational Linguistics.
\newblock \doi{10.18653/v1/2021.acl-long.231}.
\newblock URL \url{https://aclanthology.org/2021.acl-long.231}.

\bibitem[Zou et~al.(2023)Zou, Wang, Kolter, and Fredrikson]{zou2023universal}
A.~Zou, Z.~Wang, J.~Z. Kolter, and M.~Fredrikson.
\newblock Universal and transferable adversarial attacks on aligned language
  models.
\newblock \emph{arXiv preprint arXiv:2307.15043}, 2023.

\end{thebibliography}
